\documentclass[10pt,twocolumn,letterpaper]{article}

\usepackage{cvpr}              

\definecolor{cvprblue}{rgb}{0.21,0.49,0.74}
\usepackage{xcolor}    
\usepackage{colortbl}
\usepackage{arydshln}
\usepackage{tabularx}
\usepackage[pagebackref,breaklinks,colorlinks,allcolors=cvprblue]{hyperref}

\def\paperID{15693} 
\def\confName{CVPR}
\def\confYear{2025}

\title{FineVQ: Fine-Grained User Generated Content Video Quality Assessment}

\author{Huiyu Duan$^1$, Qiang Hu$^{1,*}$, Jiarui Wang$^1$, Liu Yang$^1$, Zitong Xu$^1$, Lu Liu$^1$, Xiongkuo Min$^1$, \\Chunlei Cai$^2$, Tianxiao Ye$^2$, Xiaoyun Zhang$^1$, Guangtao Zhai$^1$ \\
$^1$Shanghai Jiao Tong University, Shanghai, China\\
$^2$Bilibili Inc., Shanghai, China
\\
{\tt\small \{huiyuduan, qiang.hu, wangjiarui, ylyl.yl, lettieliu, minxiongkuo, xiaoyun.zhang,} \\
{\tt\small zhaiguangtao\}@sjtu.edu.cn, \{caichunlei, yetianxiao\}@bilibili.com} \\
}

\usepackage[export]{adjustbox} 
\renewcommand{\paragraph}[1]{\vspace{1.25mm}\noindent\textbf{#1}}

\usepackage{tabulary,array}
\newcolumntype{x}[1]{>{\centering\arraybackslash}p{#1pt}}

\usepackage{sidecap}
\usepackage{multirow}

\usepackage{pifont}

\definecolor{mred}{RGB}{238, 34, 12}
\definecolor{mgreen}{RGB}{1, 127, 0}
\definecolor{mblue}{RGB}{0, 77, 128}
\definecolor{mgreenblue}{RGB}{1,102,98}
\definecolor{orange}{RGB}{240, 120,0}

\newcommand{\bred}[1]{\textbf{\textcolor{red}{#1}}}
\newcommand{\blue}[1]{\textbf{\textcolor{mblue}{#1}}}

\newcommand{\MYhref}[3][blue]{\href{#2}{\color{#1}{#3}}} 

\begin{document}

\twocolumn[{
\vspace{-1em}
\maketitle
\centering
\vspace{-1.5em}
\includegraphics[width=1\linewidth]{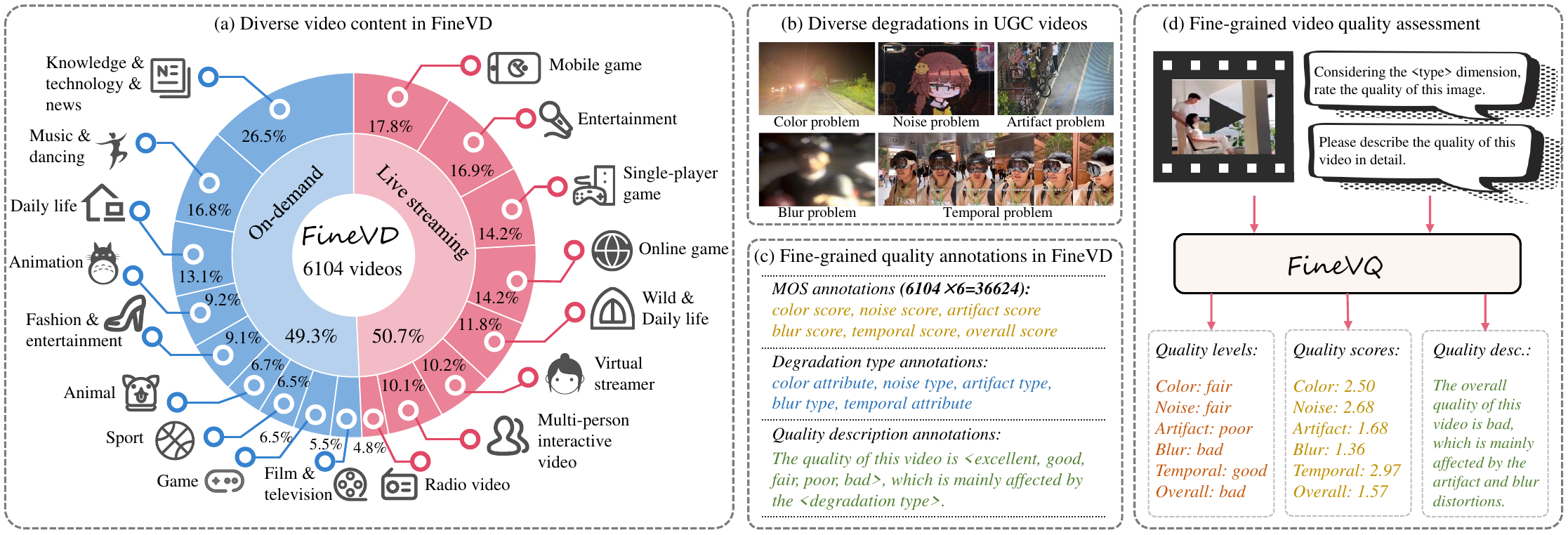}
\vspace{-2em}
\captionof{figure}{We present the fine-grained video quality assessment database and model, termed FineVD and FineVQ, respectively.
UGC videos have diverse video content but suffer from various degradation issues as shown in (a) and (b), thus it is important to provide fine-grained quality labels for subsequent video processing and recommendation tasks in addition to only providing an overall quality score.
To tackle the challenges, we construct FineVD, which includes fine-grained quality annotations for the UGC videos as shown in (c), and propose FineVQ, which has capabilities of quality rating, quality scoring, and quality attribution, as demonstrated in (d).
}
\label{fig:1_frontpage}
\vspace{1.5em}
}]
{
\renewcommand{\thefootnote}{}
\footnotetext{$^{*}$ Corresponding author.}
}


\begin{abstract}
The rapid growth of user-generated content (UGC) videos has produced an urgent need for effective video quality assessment (VQA) algorithms to monitor video quality and guide optimization and recommendation procedures.
However, current VQA models generally only give an overall rating for a UGC video, which lacks fine-grained labels for serving video processing and recommendation applications.
To address the challenges and promote the development of UGC videos, we establish the first large-scale \textbf{\underline{Fine}}-grained \textbf{\underline{V}}ideo quality assessment \textbf{\underline{D}}atabase, termed \textbf{FineVD}, which comprises 6104 UGC videos with fine-grained quality scores and descriptions across multiple dimensions.
Based on this database, we propose a \textbf{\underline{Fine}}-grained \textbf{\underline{V}}ideo \textbf{\underline{Q}}uality assessment (\textbf{FineVQ}) model to learn the fine-grained quality of UGC videos, with the capabilities of quality rating, quality scoring, and quality attribution.
Extensive experimental results demonstrate that our proposed FineVQ can produce fine-grained video-quality results and achieve state-of-the-art performance on FineVD and other commonly used UGC-VQA datasets.
Both FineVD and FineVQ are publicly available at: \MYhref[magenta]{https://github.com/IntMeGroup/FineVQ}{https://github.com/IntMeGroup/FineVQ}.
\end{abstract}

\section{Introduction}
\label{sec:intro}
Recent years have witnessed an explosion of user-generated content (UGC) videos \cite{min2024perceptual,yu2021predicting}, thanks to the evolution of video processing technologies, and the popularity of social media platforms \cite{tu2021ugc,ying2021patch}.
UGC videos cover diverse video capturing and processing conditions \cite{lu2024kvq}, thus commonly suffering from various degradations such as noise, blur, shaking, \textit{etc.}, which makes the visual quality of UGC videos vary greatly.
Understanding and predicting the quality of UGC videos has long been an important but unsolved problem, which can be applied to capture devices and social media sites to monitor video quality and guide optimization and recommendation procedures \cite{ying2021patch}.
However, different applications generally require quality labels considered from different dimensions, making a single overall quality rating insufficient for diverse downstream tasks \cite{duan2024uniprocessor,duan2022develop}.

Many UGC video quality assessment (VQA) databases have been established \cite{hosu2017konstanz,wang2019youtube,tu2021ugc,ying2021patch,wu2023towards} and numerous UGC VQA methods \cite{chen2021learning,yuan2023capturing,sun2022deep,wu2023exploring,wu2023neighbourhood,liao2022exploring,you2021long} have been proposed in the literature.
Some UGC VQA databases have focused on the in-the-wild distortions \cite{hosu2017konstanz,wang2019youtube,ying2021patch}, while several databases have considered the compression distortions in UGC videos \cite{li2020ugc,yu2021predicting,wang2021rich,zhang2023md}.
Recently, some works have also established quality assessment databases for short-form UGC (S-UGC) videos \cite{lu2024kvq}, which contain various special creation or generation modes such as special effects, numerous subtitles, \textit{etc}.
These existing databases generally only provide an overall mean quality score for a UGC video, which lacks fine-grained quality labels evaluated from multiple dimensions to facilitate subsequent diverse applications.
Some studies have also provided fine-grained labels or descriptions for UGC videos \cite{wu2023exploring,wu2023towards,QBench,you2024descriptive,you2023depicting}.
However, the description of low-level attributes is coarse-grained, and in some applications, we may still need to use specific fine-grained quality scores.

To address the fine-grained VQA problems, we establish the first large-scale \underline{Fine}-grained video quality assessment \underline{D}atabase termed \textbf{FineVD}.
Specifically, 6104 UGC videos are collected to cover a wide range of UGC scenarios, including on-demand and live-streaming applications, general and short-form video content, \textit{etc.}, as shown in Figure \ref{fig:1_frontpage}(a).
Based on the collected UGC videos, a professional team consisting of image-processing researchers is responsible for the quality labeling from six dimensions, including \textit{color, noise, artifact, blur, temporal, and overall}, in a lab environment.
As shown in Figure \ref{fig:1_frontpage}(c), FineVD includes \textit{36624} mean opinion scores (MOSs) produced from over \textit{800K} quality ratings, fine-grained degradation type labels, and quality description annotations, constituting a comprehensive quality assessment database for UGC videos.

Based on the FineVD benchmark, we further introduce the first \underline{Fine}-grained \underline{V}ideo \underline{Q}uality evaluator (\textbf{FineVQ}) to perform multi-dimensional fine-grained video quality assessment in a \textbf{\textit{one-for-all}} manner.
To achieve the comprehensive capabilities of quality rating, quality scoring, and quality depicting from multiple dimensions, FineVQ is built upon large multimodal models (LMMs) \cite{li2023blip,chen2024internvl,ye2024mplug} and leverages instruction tuning \cite{liu2024visual} and low-rank adaptation (LoRA) \cite{hulora} techniques.
Specifically, our FineVQ adopts the vision encoder used in InternVL \cite{chen2024internvl}, \textit{i.e.}, InternViT, as the spatial visual backbone, along with a motion feature extraction encoder \cite{feichtenhofer2019slowfast} as the temporal visual backbone.
A large language model, \textit{i.e.}, InternLM \cite{team2023internlm}, is used as the language backbone to facilitate the quality-related score regression and text generation.
Moreover, to extract quality-related features and refine quality-specific attributes, we impose LoRA weights on both the spatial visual backbone and the large language backbone.
Additionally, we also employ instruction tuning by inputting various question-answering pairs and training the feature projectors to enable task-specific quality awareness.
Extensive experimental results demonstrate the fine-grained quality rating, scoring, depicting abilities, and the state-of-the-art performance of FineVQ. The main highlights of this work include:

\begin{itemize}
    \item We construct the \textbf{FineVD}, the first large-scale UGC-VQA database that contains 6104 videos with corresponding over 800K subjective quality ratings and descriptions, annotated from multiple dimensions.
    \item We propose the \textbf{FineVQ}, which leverages the powerful visual representation, understanding, and description capabilities to perform quality rating, quality scoring, and quality depicting tasks.
    \item The instruction tuning and LoRA techniques are employed to enable the task-specific quality awareness in FineVQ.
    \item The extensive experimental results on FineVD and other commonly-used UGC VQA databases manifest the fine-grained VQA capabilities and state-of-the-art performance of the proposed FineVQ model.
\end{itemize}

\section{Related Work}

\paragraph{UGC-VQA Databases.}
Many VQA databases have been established in the literature to study the human perceptual quality characteristics of videos.
The early databases typically focus only on synthetic distortions, using limited source videos and manually introduced degradation types \cite{seshadrinathan2010study,de2010h,moorthy2012video,nuutinen2016cvd2014,wang2016mcl}.
In recent years, with the popularity of UGC, many UGC VQA databases have also been established considering authentic distortions in real-world or real applications.
Some UGC VQA databases \cite{nuutinen2016cvd2014,ghadiyaram2017capture,hosu2017konstanz,sinno2018large,wang2019youtube,ying2021patch} have been proposed focusing on authentic distortions, \textit{i.e.}, in-capture or in-the-wild degradations.
Several databases \cite{li2020ugc,yu2021predicting,wang2021rich,zhang2023md,icme21} have been presented considering both synthetic and authentic distortions.
Moreover, considering the UGC videos in the aforementioned databases are sourced from general media platforms (\textit{e.g.}, Youtube), a recent database, KVQ \cite{lu2024kvq}, also contributes a UGC VQA database specifically focused on short-form videos.
In contrast, our FineVD includes both on-demand and live-streaming videos, as well as general UGC and short-form UGC formats.
And fine-grained labels in FineVD can lead to more applications compared to an overall score.

\begin{figure*}[t]\centering
\vspace{-0.2em}
\includegraphics[width=1\linewidth]{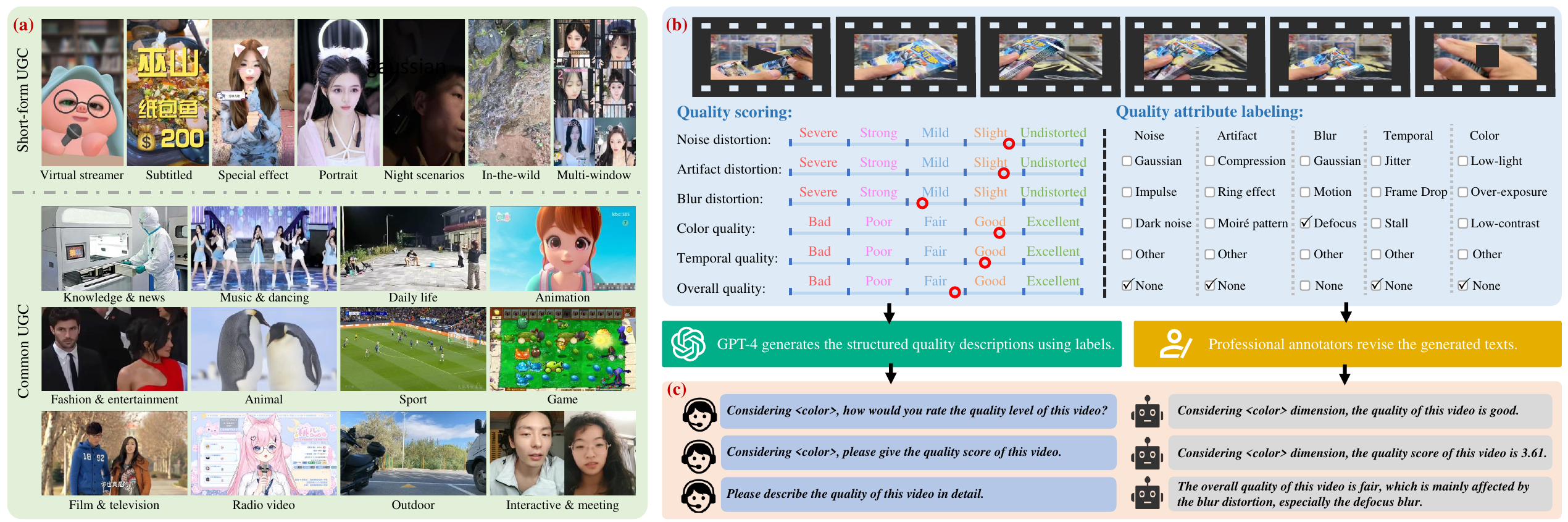}
\vspace{-1.9em}
\caption{An overview of the content and construction process of FineVD. 
(a) Example videos from our database, which contains both common UGC videos and short-form UGC videos. 
(b) The illustration of subjective data annotation methods, including both quality scoring and quality attribute labeling processes.
(c) The quality-related question-answering pairs generated by GPT-4 and revised by human annotators.
}
\vspace{-1.6em}
\label{fig:2}
\end{figure*}

\paragraph{UGC-VQA Models.}
With the establishment of numerous UGC VQA databases, a variety of UGC-VQA models have also been proposed in the literature.
Traditional methods \cite{kancharla2021completely,korhonen2019two,mittal2015completely,saad2014blind,zheng2022completely} generally extract and regress handcrafted features to predict the video quality, which lack adaptability to complex UGC scenarios.
With the development of deep learning, many deep neural network-based (DNN-based) VQA models have been presented and have achieved superior performance.
A large number of VQA methods \cite{li2019quality,chen2021learning,yuan2023capturing,sun2022deep,wu2022fast,wu2023exploring,wu2023neighbourhood,liao2022exploring,you2021long,ge2024lmm} employ pre-trained DNN models to extract semantic features, and train quality regressors to predict quality scores.
Some studies \cite{wu2021no,liu2021spatiotemporal,chen2021unsupervised,chen2021contrastive,mitra2022multiview} have also explored using self-supervised learning strategies to train VQA models.
However, these methods lack distortion understanding ability and can only predict an overall quality score for a UGC video.

\paragraph{Large Multimodal Models.}
The rapid evolution of large language models \cite{achiam2023gpt,touvron2023llama,wei2022chain} and vision language pre-training techniques \cite{radford2021learning,li2022blip} has explored the advancement of large multimodal models \cite{liu2024visual,liu2024improved,lu2023empirical,zhuminigpt,li2023blip,chen2024internvl,ye2024mplug}, which has superior multimodal joint understanding and interpretation abilities.
Based on LMMs, some works have also explored the ability of the LMMs on low-level visual understanding \cite{chen2023x,QBench,QAlign,Qinstruct}, which can also effectively predict and describe the quality levels.
However, precise visual quality assessment, \textit{i.e.,} MOS prediction, is still useful in real applications, and these models lack abilities to give fine-grained quality levels and scores from multi-dimensions.

\section{The FineVD Database}

In this section, we introduce the proposed large-scale fine-grained video quality assessment database (\textbf{FineVD}).
The database includes 6104 UGC videos, with corresponding 805728 human opinion ratings.
Our FineVD exhibits the advantages of comprehensive UGC content and multi-dimensional labeling, which can significantly facilitate the advancement of UGC research.

\subsection{Video Collection}
The video collection process of FineVD follows two key principles: (1) covering a wide range of UGC scenarios and diverse distortions as comprehensively as possible, and (2) reflecting practical online statistics and applications of popular video platforms.
To this end, we first collect a large-scale video dataset from the popular UGC video platform Bilibili \cite{bilibili}.
Then we manually screen the collected videos to ensure the diversity of video scenarios and quality attributes.
Finally, a total of 6104 UGC videos are included in our database.
As shown in Figure \ref{fig:1_frontpage}, FineVD contains both on-demand videos and live-streaming videos, in which the on-demand UGC scenes consist of knowledge\&technology\&news, music\&dancing, daily life, animation, fashion\&enterainment, animal, sport, \textit{etc.}, while live-streaming videos include scenes of mobile games, entertainment, single-player games, online games, wild, virtual streamer, \textit{etc.}
Figure \ref{fig:2} demonstrates that our FineVD contains both common UGC scenarios as aforementioned, and short-form UGC contents \cite{lu2024kvq} such as virtual streamer, subtitle, special effect, multi-window interaction, \textit{etc.}, which further manifest the content diversity of the established database.
Moreover, Figure \ref{fig:1_frontpage}(b) also shows that FineVD covers a variety of degradations.

\begin{figure*}[t]\centering
\vspace{-0.2em}
\includegraphics[width=1\linewidth]{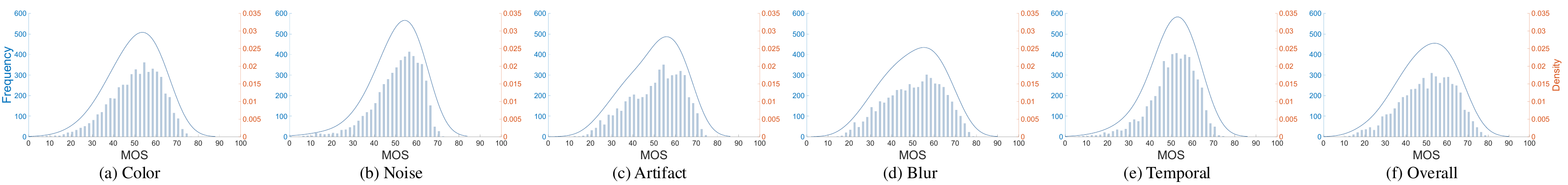}
\vspace{-2em}
\caption{The MOS distribution of FineVD in terms of different perspectives, \textit{i.e.}, color, noise, artifact, blur, temporal, and overall.
}
\vspace{-1.3em}
\label{fig:3}
\end{figure*}
\begin{figure}[t]\centering
\includegraphics[width=1\linewidth]{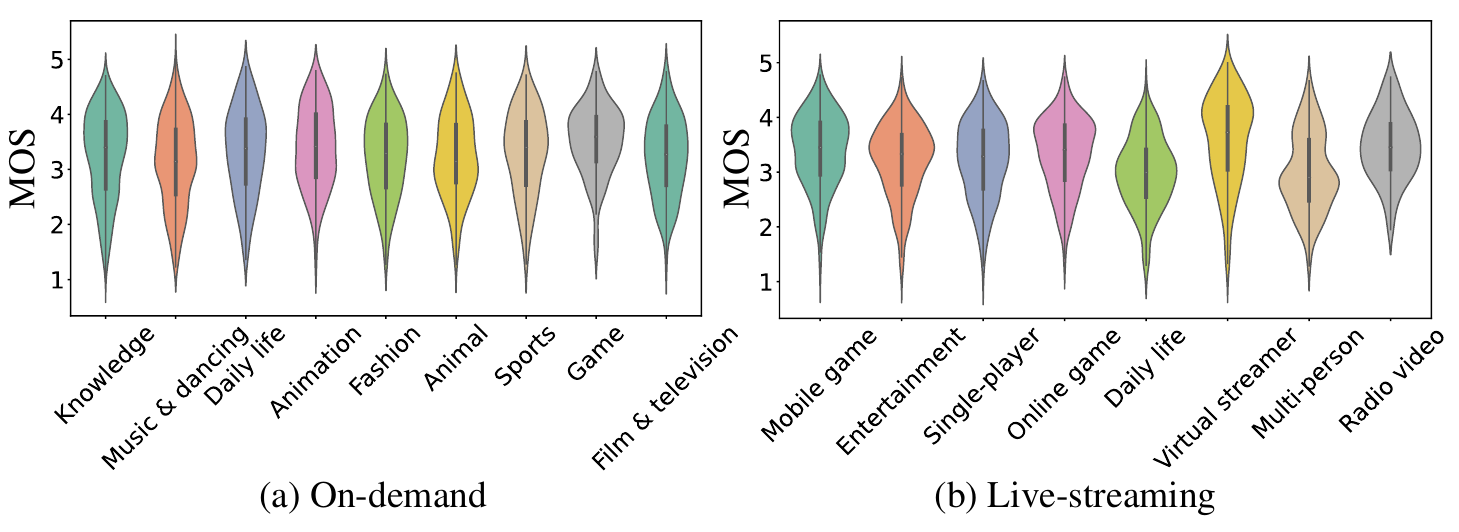}
\vspace{-1.9em}
\caption{The MOS distribution of \textit{``overall score''} in terms of different video contents.
(a) MOS distribution for on-demand videos. (b) MOS distribution for live-streaming videos. 
}
\vspace{-1em}
\label{fig:4}
\end{figure}

\subsection{In-lab Subjective Study}
Different from the crowd-sourced method adopted by some previous UGC VQA databases, such as LSVQ \cite{ying2021patch}, our subjective study is carried out in a lab environment.
A total of 22 professional annotators participate in the experiments.
The videos are displayed on a DELL UltraSharp monitor with a resolution of 3840$\times$2160, and the viewers are seated at a distance of about 2 feet from the monitor.
Before the subjective experiments, we provide clear rating criteria with numerous examples to train the subjects.
As shown in Figure \ref{fig:2}(b), the subjects are introduced to evaluate the quality of a video from six perspectives, including \textit{noise, artifact, blur, color, temporal, and overall}.
For the quality rating task, subjects are asked to give their opinions on the videos in five categories from six perspectives.
In particular, for \textit{noise, artifact, and blur} distortions, the five-category levels are introduced as severe, strong, mild, slight, and undistorted, respectively, while for \textit{color, temporal, and overall} perspectives, the five-category levels are explained as bad, poor, fair, good, and excellent, respectively.
Besides the quality scores, identifying the distortion or quality attributes is also important for applications such as video processing, video recommendation, \textit{etc.}
Thus, we also collect the quality attribute labels as shown in Figure \ref{fig:2}(b).
For each distortion category, we provide three commonly encountered distortion types and two additional options ``other'' and ``none''.
In particular, for the ``other'' option, the subjects can manually enter the quality attributes using a jump window.

Based on the collected quality scores and attribute labels, we further prepare language question-answering (QA) pairs for tuning LMMs \cite{QAlign,you2023depicting,you2024descriptive}.
Specifically, we first generate QA pairs automatically using an excellent large language model, \textit{i.e.,} GPT-4 \cite{achiam2023gpt}, and then we revise the generated texts manually to avoid ambiguous QA pairs.
Finally, we introduce three tasks including quality-level prediction, quality score regression, and quality description, respectively, as shown in Figure \ref{fig:2}(c).
The generated QA pairs can significantly facilitate the subsequent training for our FineVQ model.

\subsection{Subjective Data Analysis}
We follow the suggestions given in \cite{series2012methodology} to conduct the outlier detection and subject rejection.
As a result, no subject is rejected, and each image is rated by 22 valid subjects.
Among all scores, about 3.16\% of the total subjective evaluations are identified as outliers and are subsequently removed.
For the remaining valid subjective ratings, we convert these raw ratings into Z-scores \cite{duan2022confusing} as follows:
\vspace{-0.5em}
\begin{equation}
  z_{ij}=\frac{m_{ij}-\mu_{i}}{\sigma_{i}},
\vspace{-0.5em}
\end{equation}
where $m_{ij}$ is the raw rating given by the $i$-th subject to the $j$-th image, $\mu_{i}$ is the mean rating given by subject $i$, $\sigma_{i}$ is the standard deviation.
Then these scores are linearly scaled to the range $[0, 100]$ and obtain the rescaled z-scores $z_{ij}'$, which are averaged over subjects to obtain the final mean opinion scores (MOSs):
\vspace{-0.5em}
\begin{equation}
  \text{MOS}_{j}=\frac{1}{N}\sum_{i=1}^{N}z_{ij}',
\vspace{-0.5em}
\end{equation}
where $N$ is the total number of subjects.

Based on the MOSs, we further analyze the collected data.
Specifically, we first visualize the MOS distribution for six evaluation dimensions, respectively, as shown in Figure \ref{fig:3}.
It can be observed that all MOS distributions of six dimensions cover a wide range.
Moreover, although the distributions from different evaluation perspectives are generally similar, there are distinct differences in the details.
To investigate the MOS distribution differences in terms of different content and applications, we further visualize the MOS distribution for 9 on-demand categories and 8 live-streaming categories in Figure \ref{fig:4}, respectively.
We can observe that the MOS distributions of different contents are generally similar, except for the game content in on-demand videos and virtual streamer and multi-person interaction categories for live-streaming videos.
This manifests that these video contents are unusual compared to other UGC videos.
Comparing Figure \ref{fig:4}(a)\&(b), we can also observe that the MOS distributions are significantly different for on-demand and live-streaming videos.
Additionally, the extensive coverage of all video categories also demonstrates the abundance of our database.

\begin{figure*}[t]\centering
\vspace{-0.3em}
\includegraphics[width=1\linewidth]{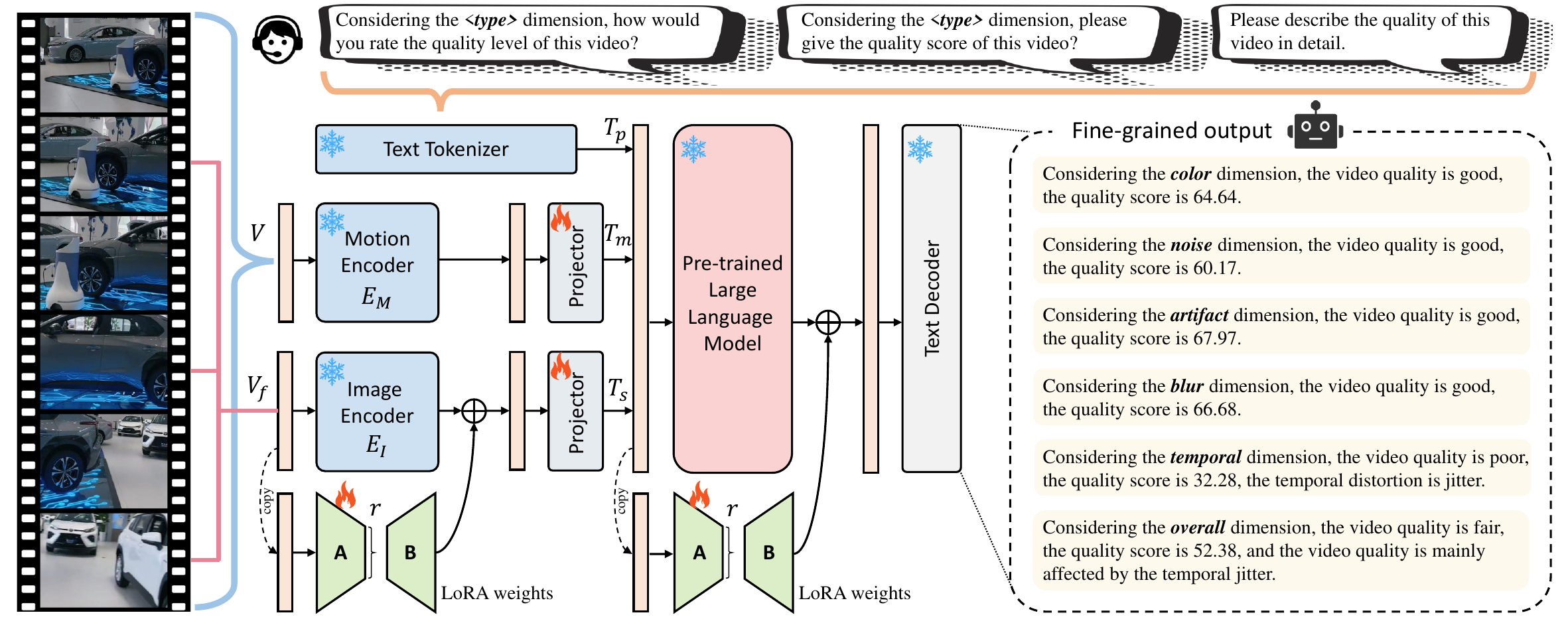}
\vspace{-1.9em}
\caption{An overview of our proposed FineVQ model. Our model consists of three feature encoders, including an image feature extractor for extracting spatial features from sparse video frames, a motion feature extractor for extracting motion features from the entire video, and a text encoder for extracting aligned text features from prompts. The extracted features are then aligned through projectors and fed into a pre-trained LLM to generate the output results. LoRA weights are introduced to the pre-trained image encoder and the large language model to adapt the models to the quality assessment task.
}
\vspace{-1.2em}
\label{fig:5}
\end{figure*}

\section{The FineVQ Approach}

In this section, we introduce our \textbf{\textit{one-for-all}} video quality assessment method, \textbf{FineVQ}, towards giving quality levels, predicting quality scores, and depicting quality attributes from multiple perspectives using one model.

\subsection{Overall Architecture}
The overall framework of FineVQ is depicted in Figure \ref{fig:5}, which takes both UGC videos and user prompts as input, and outputs quality-related answers according to both questions and videos.
FineVQ begins with extracting visual features and text features from UGC videos and user prompts respectively.
For the visual feature extraction part, two vision backbones are introduced.
An image encoder is used to extract spatial features from selected video frames, and a motion encoder is applied to extract motion features from the entire video.
Then two projectors $P_I$ and $P_M$ are used to project the extracted features into language space and output tokens $T_s$ and $T_m$.
For text feature extraction, a text tokenizer is utilized to encode the user prompt into tokens $T_p$.
The extracted three feature tokens $T_s$, $T_m$, and $T_p$ are concatenated and input into a pre-trained large language model to generate the final outputs.

\subsection{Model Design}

\paragraph{Visual Encoding.} The visual encoding part involves two visual feature extractors, \textit{i.e.,} an image encoder and a motion encoder.
Specifically, for an input UGC video $V$, we uniformly collect 8 frames $V_f$ from the video and then extract the features from these frame images using the image encoder.
The image encoder $E_I$ is built on a pre-trained vision transformer (ViT), \textit{i.e.}, InternViT \cite{chen2024internvl}, which is pre-trained on the LAION-en dataset \cite{schuhmann2022laion} using text-image contrastive learning.
To align the extracted features with the input space of the large language model, a projector $P_I$ with two multilayer perceptron (MLP) layers is applied.
The process can be formulated as:
\vspace{-0.5em}
\begin{equation}
    T_s = P_I(E_I(V_f)),
\vspace{-0.5em}
\end{equation}
where $T_s$ is the mapped frame feature tokens.
Considering the sparse frame can not include meticulous temporal features, a motion feature extractor $E_M$ is also introduced to extract slow-fast motion features from a video \cite{feichtenhofer2019slowfast}.
Similarly, a projector $P_M$ with two MLP layers is also introduced to ensure the feature dimension consistency and semantic feature alignment between the motion features and LLM input.
The process can be formulated as:
\vspace{-0.5em}
\begin{equation}
    T_m = P_M(E_M(V)),
\vspace{-0.5em}
\end{equation}
where $T_m$ is the mapped moton feature tokens.

\begin{table*}
\vspace{-0.3em}
\setlength{\belowcaptionskip}{-0.01cm}
\centering
\belowrulesep=0pt
\aboverulesep=0pt
\renewcommand\arraystretch{1.2}
\caption{Performance of state-of-the-art models and the proposed FineVQ on our established FineVD database in terms of the \textit{quality scoring task}. $\spadesuit$, $\diamondsuit$, and $\heartsuit$ denote the traditional IQA methods, traditional VQA methods, and DNN-based VQA metrics, respectively. It should be noted that the DNN-based models are trained with separate weights for different dimensions, while FineVQ complies with a \textbf{\textit{one-for-all}} fashion. The best results are highlighted in \bred{red}, and the second-best results are highlighted in \blue{blue}.
}
\vspace{-0.8em}
   \resizebox{\linewidth}{!}{\begin{tabular}{lcccccccccccccccccc}
    \toprule[1pt]
    Dimension  &\multicolumn{3}{c}{Color}&\multicolumn{3}{c}{Noise}&\multicolumn{3}{c}{Artifact}&\multicolumn{3}{c}{Blur}&\multicolumn{3}{c}{Temporal}&\multicolumn{3}{c}{Overall}\\
  \cmidrule(lr){2-4} \cmidrule(lr){5-7} \cmidrule(lr){8-10} \cmidrule(lr){11-13} \cmidrule(lr){14-16} \cmidrule(lr){17-19}
 Methods / Metrics
&SRCC$\uparrow$&KRCC$\uparrow$&PLCC$\uparrow$&SRCC$\uparrow$&KRCC$\uparrow$&PLCC$\uparrow$&SRCC$\uparrow$&KRCC$\uparrow$&PLCC$\uparrow$&SRCC$\uparrow$&KRCC$\uparrow$&PLCC$\uparrow$&SRCC$\uparrow$&KRCC$\uparrow$&PLCC$\uparrow$&SRCC$\uparrow$&KRCC$\uparrow$&PLCC$\uparrow$\\
    \midrule
    $\spadesuit$NIQE~\cite{niqe}&0.3273 &0.2246 &0.2368 &0.2682 &0.1873 &0.1417 &0.3006&0.2106&0.1649&0.3236&0.2259&0.1870&0.2777&0.1927&0.1630&0.3192&0.2222 &0.2019 \\
    $\spadesuit$QAC~\cite{qac}&0.1160 &0.0852 &0.0789 &0.0072 &0.0559 &0.0067 &0.0549 &0.0154 &0.0393 &0.0862 &0.0146 &0.0600 &0.0410 &0.0103 &0.0280 &0.0739 &0.0180 &0.0512 \\
    $\spadesuit$HOSA~\cite{hosa}&0.3593&0.2449&0.3233&0.2937&0.2440&0.2385&0.3457&0.2419&0.3067&0.3878&0.2719&0.3451&0.2983&0.2078&0.2429&0.3627&0.2525&0.3236 \\ \hdashline
    $\diamondsuit$TLVQM~\cite{korhonen2019two}&0.6346&0.4550&0.6417&0.6357&0.4543&0.5909&0.6608&0.4756&0.6526&0.6639&0.4768&0.6588&0.6816&0.4968&0.6996&0.6537&0.4698&0.6546\\
    $\diamondsuit$VIDEVAL~\cite{tu2021ugc}
    &0.6922&0.5034&0.6943&0.6912&0.5033&0.6514&0.7440&0.5500&0.7370&0.7610&0.5649 &0.7637&0.7174&0.5305&0.7123&0.7310&0.5373&0.7307\\
    $\diamondsuit$RAPIQUE~\cite{tu2021rapique}
    &0.6203&0.4470&0.6330&0.6048&0.4284&0.5607&0.6258&0.4484&0.6276&0.6572&0.4778&0.6717&0.5554&0.3945&0.5429&0.6379&0.4597&0.6501 \\ \hdashline
    $\heartsuit$VSFA~\cite{li2019quality}&0.7617 &0.5673 &0.7837 &0.7635 &0.5707 &0.7278 &0.8006 &0.6081&0.8170 &0.7772 &0.5858 &0.8001 &0.7276 &0.5355 &0.7018 &0.7730 &0.5825 &0.7929\\
    $\heartsuit$GSTVQA~\cite{chen2021learning}&0.7747&0.5831 &0.7761 &0.7883 &0.5974 &0.7448 &0.8121 &0.6200 &0.8187 &0.8101 &0.6185 &0.8202 &0.7533 &0.5593 &0.7093 &0.7834 &0.5906 &0.7825 \\
    $\heartsuit$SimpleVQA~\cite{sun2022deep}&0.8086 &0.6097 &0.8058 &0.8070 &0.6133 &0.7634 &\blue{0.8465} &\blue{0.6550} &\blue{0.8487} &\blue{0.8466} &\blue{0.6558} &\blue{0.8519} &\blue{0.7746} &\blue{0.5744} &0.7417 &0.8311 &0.6401 &0.8358 \\
    $\heartsuit$FAST-VQA~\cite{wu2022fast}&0.8017&0.6078&0.8183&\blue{0.8093}&\blue{0.6209}&	\blue{0.7758}&0.8176&0.6295&0.8328&0.8352&0.6500&0.8513&0.7560&0.5609&0.7393&0.8348&	0.6476&\blue{0.8474}\\
    $\heartsuit$DOVER~\cite{wu2023exploring}&\blue{0.8244}&\blue{0.6311}&\blue{0.8311}&0.8018&0.6055&0.7424&0.8265&0.6338&0.8289&0.8404&0.6504&0.8355&0.7664&0.5700&\blue{0.7569}&\blue{0.8422}&\blue{0.6517}&0.8393\\
   \rowcolor{gray!20} \textbf{FineVQ (Ours)} &\bred{0.8495}&\bred{0.6665}&\bred{0.8527}&\bred{0.8444}&\bred{0.6613}&\bred{0.7986}&\bred{0.8852}&\bred{0.7111}&\bred{0.8921}&\bred{0.8711}&\bred{0.6957}&\bred{0.8833}&\bred{0.8085}&\bred{0.6171}&\bred{0.7597}&\bred{0.8834}&\bred{0.7118}&\bred{0.8891}\\
    \bottomrule[1pt]
  \end{tabular}}
  \label{tab1_finevd}
  \vspace{-1.2em}
\end{table*}
\begin{table}
\setlength{\belowcaptionskip}{-0.01cm}
\centering
\belowrulesep=0pt
\aboverulesep=0pt
\renewcommand\arraystretch{1.2}
\caption{Performance of state-of-the-art LMMs and the proposed FineVQ on our established FineVD database in terms of the \textit{quality attribute prediction task}. The \textit{``yes-or-no''} type represents the judgment on whether the corresponding dimension is degraded. The \textit{``which''} type indicates which distortion exists or has the most impact on the quality of the video.
}
\vspace{-0.8em}
   \resizebox{\linewidth}{!}{\begin{tabular}{lccccccc}
    \toprule[1pt]
    Question Type&\multicolumn{5}{c}{\textit{Yes-or-no}}&\multicolumn{2}{c}{\textit{Which}}\\
  \cmidrule(lr){2-6} \cmidrule(lr){7-8}
 Model / Attribute&Color&Noise&Artifact&Blur&Temporal&Exist&Most\\
    \midrule
    VideoLLaMA2 (\textit{7B}) ~\cite{damonlpsg2024videollama2}&57.87\%&63.98\%&51.87\%&61.91\%&\blue{86.42\%}&25.59\%&28.44\% \\
    Video-LLaVA (\textit{7B}) ~\cite{lin2023video}&26.48\%&27.26\%&43.90\%&64.67\%&13.09\%&16.73\%&36.71\%\\
    VideoChat2 (\textit{7B}) ~\cite{li2024mvbench}&18.25\%&18.80\%&24.26\%&31.15\%&15.19\%&10.93\%&13.99\% \\
    Video-ChatGPT (\textit{7B}) ~\cite{Maaz2023VideoChatGPT}&28.35\%&27.66\%&44.49\%&62.89\%& 18.60\%&20.37\%&24.11\% \\
    InternVL2 (\textit{8B}) ~\cite{chen2024internvl}&58.46\%&63.58\%&50.69\%&54.33\% &70.28\%&\blue{28.25\%}&\blue{43.21\%} \\
    MiniCPM-V (\textit{8B}) ~\cite{yao2024minicpm}&70.47\%&\blue{65.85}\%&\bred{52.66\%}&46.26\%&71.95\%&25.98\% &30.81\% \\
    Qwen2-VL (\textit{7B}) ~\cite{Qwen2VL}&\blue{71.16\%}&44.78\%&42.91\%&41.14\%&54.33\%&23.33\%&27.85\% \\
    LLaVA-NeXT (\textit{7B}) ~\cite{li2024llavanext-ablations}&26.87\%&27.26\%&43.90\%&\bred{64.76\%}&13.09\%&17.81\%&37.89\% \\
   \rowcolor{gray!20} \textbf{FineVQ (Ours)} &\bred{73.52\%}&\bred{72.74\%}&\blue{51.87\%}&\bred{64.76\%}&\bred{86.91\%}&\bred{91.93\%}&\bred{65.06\%}\\
    \bottomrule[1pt]
  \end{tabular}}
  \label{tab2_llm}
\vspace{-1.3em}
\end{table}

\paragraph{Feature Fusion via the LLM.}
For a given prompt $P$, it is first encoded into the text tokens $T_p$ via a text tokenizer.
Then the text tokens $T_p$ are concatenated with the well-aligned visual tokens $T_s$ and $T_m$ as the input to the LLM.
In particular, InternLM-8B \cite{team2023internlm} is used as the pre-trained LLM to combine visual tokens and text tokens to perform multimodal learning.
Finally, the output features of the LLM are decoded with a text decoder and then projected to the prompt-corresponded quality space.

\subsection{Instruction Tuning and LoRA Adaptation}

\paragraph{Instruction Tuning.} It is of great significance to achieve \textbf{one-for-all} VQA, which is conducive to multi-dimensional assessment.
However, previous VQA models generally need to train multiple weights for different dimensions respectively.
Benefiting from the generalization ability of LLMs, many recent works have verified the effectiveness of using the instruction tuning strategy for one-for-all tasks \cite{liu2024visual,liu2024improved}.
As shown in Figure \ref{fig:5}, our input prompts contain various question types from multiple dimensions.
For different instructions, FineVQ trains visual feature projectors to tune the LLM to align textual and visual semantics for joint reasoning.
As a result, FineVQ can evaluate the video quality from multiple dimensions using one model weight.

\paragraph{LoRA Adaptation.}
Fine-tuning LLMs is generally resource consuming but can lead to better performance.
To further improve the performance of FineVQ, we adopt the LoRA technique \cite{hulora} to fine-tune the model.
Specifically, we employ LoRA to the image encoder $E_I$ and the LLM.
LoRA models the changes $\Delta \textbf{W}\in\mathbb{R}^{n\times m}$ for each layer's weights $\textbf{W}\in\mathbb{R}^{n\times m}$ as $\Delta \textbf{W}=\textbf{A}\textbf{B}$, where $\textbf{A}\in\mathbb{R}^{n\times r}$ and $\textbf{B}\in\mathbb{R}^{r\times m}$.
The rank is constrained by $r\ll\{n,m\}$ to achieve parameter efficiency.
Given the original output $\textbf{h}=\textbf{W}\textbf{x}$, the forward pass of LoRA is:
\vspace{-0.5em}
\begin{equation}
    \textbf{h}=\textbf{W}\textbf{x}+\Delta \textbf{W}\textbf{x}=(\textbf{W}+\textbf{A}\textbf{B})\textbf{x}.
\vspace{-0.5em}
\end{equation}
With the LoRA, FineVQ can effectively adapt to the fine-grained VQA task.

\begin{table*}[t]
\vspace{-0.3em}
\setlength{\belowcaptionskip}{-0.01cm}
\centering
\belowrulesep=0pt
\aboverulesep=0pt
\renewcommand\arraystretch{1.15}
  \scriptsize
  \caption{Performance comparison between state-of-the-art VQA methods and the proposed FineVQ on six UGC VQA databases. The ``N/A'' means missing results in the original paper. The best results and the second-best results are highlighted in \bred{red} and \blue{blue}, respectively.
  }
\vspace{-0.9em}
\resizebox{0.8\textwidth}{!}{
  \begin{tabular}{lcccccccccccc}
    \toprule
    \multirow{2}{*}{Method} & \multicolumn{2}{c}{LIVE-YT-Gaming~\cite{yu2023subjective}} & \multicolumn{2}{c}{KoNViD-1k~\cite{hosu2017konstanz}} & \multicolumn{2}{c}{YouTube-UGC~\cite{wang2019youtube}} & \multicolumn{2}{c}{LIVE-VQC~\cite{sinno2018large}} & \multicolumn{2}{c}{LSVQ$_\text{test}$~\cite{ying2021patch}} & \multicolumn{2}{c}{LSVQ$_\text{1080p}$~\cite{ying2021patch}}\\
  \cmidrule(lr){2-3} \cmidrule(lr){4-5} \cmidrule(lr){6-7} \cmidrule(lr){8-9} \cmidrule(lr){10-11} \cmidrule(lr){12-13}
    & SRCC$\uparrow$ & PLCC$\uparrow$ & SRCC$\uparrow$ & PLCC$\uparrow$ & SRCC$\uparrow$& PLCC$\uparrow$ & SRCC$\uparrow$ & PLCC$\uparrow$ & SRCC$\uparrow$ & PLCC$\uparrow$ & SRCC$\uparrow$ & PLCC$\uparrow$ \\ 
    \midrule[0.3pt]
    VIQE~\cite{zheng2022completely} &N/A &N/A &0.628&0.638
     &0.513 &0.476  &0.659 &0.694 &N/A &N/A &N/A &N/A \\
    TLVQM~\cite{korhonen2019two}  &  0.748& 0.756  & 0.773  & 0.768    & 0.669  & 0.659 &0.798 & 0.802& 0.772  & 0.774   & 0.589  & 0.616 \\
    RAPIQUE~\cite{tu2021rapique} &0.771&0.815&0.803&0.817 & 0.759&0.768 &0.754&0.786 &N/A&N/A &N/A &N/A
    \\
    VIDEVAL~\cite{tu2021ugc}   & 0.807   & 0.812 & 0.773& 0.768& 0.669& 0.659& 0.752&0.751 & 0.795& 0.783& 0.545& 0.554\\ \hdashline
    VSFA~\cite{li2019quality}  &  0.776  &  0.801 & 0.773& 0.775& 0.724& 0.743&0.773&0.795 & 0.801& 0.796& 0.675& 0.704          \\
    GSTVQA~\cite{chen2021learning}  &N/A &N/A & 0.814& 0.825 & N/A  & N/A &0.788&0.796 & N/A         & N/A &N/A &   N/A       \\
    PVQ~\cite{ying2021patch} &N/A &N/A & 0.791 & 0.786 &N/A &N/A &  0.827&0.837  & 0.827&0.828 &0.711&0.739 \\ 
    SimpleVQA~\cite{sun2022deep} & 0.861  & 0.866 & 0.856& 0.860  & 0.847          & 0.856 & N/A&N/A & 0.867 & 0.861 & 0.764 & 0.803          \\ 
    FastVQA~\cite{wu2022fast}   &\blue{0.869}  &0.880   & 0.891 & 0.892 & 0.855          & 0.852&0.849 &0.862  & 0.876  & 0.877 & 0.779  & 0.814          \\ 
    Dover~\cite{wu2023exploring}   &0.852  &0.868   &0.908 & 0.910  & 0.841& 0.851&0.844& 0.875 &0.877  &0.878 &0.778  &0.812         \\ 
    MinimalisticVQA~\cite{sun2024analysis} & 0.857 & \blue{0.888} & 0.889 & 0.890 & 0.890 & 0.891 & 0.842 & 0.854 & 0.881 & 0.879 & 0.781 & 0.820 \\
    KSVQE~\cite{lu2024kvq} & N/A & N/A & \bred{0.922} & \bred{0.921} & \blue{0.900} & \blue{0.912} &\blue{0.861}&\blue{0.883} & \blue{0.886} & \blue{0.888} & \blue{0.790} & \blue{0.823}  \\
    \rowcolor{gray!20} \textbf{FineVQ (Ours)} & \bred{0.912} & \bred{0.926} & \blue{0.915} & \blue{0.910} & \bred{0.910} & \bred{0.914} &\bred{0.895}&\bred{0.895} & \bred{0.900} & \bred{0.900} & \bred{0.828} & \bred{0.857}  \\
     \bottomrule
  \end{tabular}}
  \label{tab3_sota}
\vspace{-1.3em}
\end{table*}

\section{Experimental Evaluation}

\subsection{Implementation Details}

For the image encoder, we use a pre-trained InternViT \cite{chen2024internvl} as the backbone and uniformly extract 8 frames from the video as the input.
The frame images are resized to $448 \times 448$ and then input.
For the motion extractor, we use the slow-fast \cite{feichtenhofer2019slowfast} as the backbone and input the entire video frames.
The video frames are also resized to $448 \times 448$ as the input.
The LLM used in FineVQ is InternLM-8B \cite{team2023internlm}, and the input token channel dimension is 4096.
The image encoder, motion encoder, and LLM are all frozen during training.
The rank parameter of LoRA \cite{hulora} in FineVQ is set to 16.
The model is trained on 4 NVIDIA RTX A6000 GPUs and flash-attn \cite{dao2022flashattention,daoflashattention} is used to save the GPU memory.
We use AdamW Optimizer to train the network for 10 epochs with a batch size of 8.
The initial learning rate is $2e^{-4}$ and gradually reduces to $1e^{-6}$ with cosine annealing.

\subsection{Evaluation on the FineVD}

\paragraph{Experimental Settings.} 
We first follow the previous VQA research settings \cite{ying2021patch,lu2024kvq} and split the UGC videos into training, validation, and test sets with a ratio of 4:1:1.
A total of 11 state-of-the-art VQA models are selected as benchmark methods in our FineVD, which can be categories into three groups: (1) traditional image quality assessment (IQA) methods, including NIQE \cite{niqe}, QAC \cite{qac}, HOSA \cite{hosa}; (2) traditional video quality assessment (VQA) methods, including TLVQM \cite{korhonen2019two}, VIDEVAL \cite{tu2021ugc}, RAPIQUE~\cite{tu2021rapique}; (3) DNN-based VQA methods, including VSFA~\cite{li2019quality}, GSTVQA~\cite{chen2021learning}, SimpleVQA~\cite{sun2022deep}, FAST-VQA~\cite{wu2022fast}, DOVER~\cite{wu2023exploring}.
The DNN-based VQA models are trained using their official codes with the default settings.
We train these models separately for different dimensions in FineVD.
Three evaluation metrics, including Spearman rank-order correlation coefficient (SRCC), Kendall rank-order correlation coefficient (KRCC), and Pearson linear correlation coefficient (PLCC), are adopted to measure the performance of VQA models.

For the quality attribute prediction task, we use the same dataset split setting as mentioned above. 
A total of 7 large multimodal models that accept video as input are selected as benchmark methods, including VideoLLaMA2~\cite{damonlpsg2024videollama2}, Video-LLaVA~\cite{lin2023video}, VideoChat2~\cite{li2024mvbench}, Video-ChatGPT~\cite{Maaz2023VideoChatGPT}, InternVL2~\cite{chen2024internvl}, MiniCPM-V~\cite{yao2024minicpm}, Qwen2-VL~\cite{Qwen2VL}, LLaVA-NeXT~\cite{li2024llavanext-ablations}.
We set two types of questions including \textit{``yes-or-no''} and \textit{``which''} to test the ability of LMMs and our FineVQ on quality-related low-level vision tasks \cite{QBench}.
Specifically, for ``yes-or-no'' questions, we query the LMMs with the prompts of \textit{``Is these any $<$dimension$>$ distortion in this video?''}, while for ``which'' questions, we infer the LMMs with the prompts of \textit{``Which distortion exists in this video?''} and \textit{``Which distortion has the most impact on the quality of this video?''}
We adopt prediction accuracy as the metric for evaluating the performance of the quality attribute prediction task.

\begin{table}

\centering

\caption{The cross-dataset evaluation results. The models are trained on other datasets and tested on our FineVD.
}
\vspace{-2mm} 
\centering

\resizebox{\linewidth}{!}{\begin{tabular}{lcccccc}
   \toprule
\textit{Train on:} & \multicolumn{2}{c}{KoNViD-1k~\cite{hosu2017konstanz}} & \multicolumn{2}{c}{YouTtbe-UGC~\cite{wang2019youtube}}  & \multicolumn{2}{c}{LIVE-VQC~\cite{sinno2018large}}\\  \cmidrule(lr){2-3} \cmidrule(lr){4-5} \cmidrule(lr){6-7}
 \textit{Test on:} FineVD   & SRCC$\uparrow$&PLCC$\uparrow$ & SRCC$\uparrow$&PLCC$\uparrow$ & SRCC$\uparrow$&PLCC$\uparrow$\\  
     \midrule
   SimpleVQA~\cite{sun2022deep}  &0.5947&0.6330 &0.6050&0.6265 & 0.4631& 0.4753  \\
   FastVQA~\cite{wu2022fast}   &0.5155&0.5155   &0.5990&0.6454 &0.5730&0.5883   \\
   \rowcolor{gray!20} FineVQ (Ours)  & \textbf{0.6475}&\textbf{0.7020}& \textbf{0.6261}&\textbf{0.6520}&\textbf{0.6125}&\textbf{0.6550}  \\
    \bottomrule
  \end{tabular}
  } 
  \vspace{-1.5em}
  \label{tab4_cross1}
\end{table}

\paragraph{Results on The Quality Scoring Task.}
We first evaluate the performance of our FineVQ and state-of-the-art VQA models on the established FineVD dataset, as shown in Table \ref{tab1_finevd}.
We first observe that the traditional IQA models typically perform poorly on FineVD, especially for the temporal dimension.
Traditional VQA models can generate meaningful quality scores, but are still far from satisfactory.
DNN-based models generally achieve better results than traditional VQA and IQA methods, and our FineVQ consistently achieves the best performance in terms of all dimensions and all metrics.
By comparing the performance of FineVQ across different dimensions, it can be observed that the proposed model performs relatively worse on the temporal dimension compared to other dimensions.
This indicates that although we have incorporated the motion features into FineVQ, the temporal features are still insufficient and can be further improved.

\paragraph{Results on The Quality Attribute Prediction Task.}
Table \ref{tab2_llm} shows the results of the state-of-the-art large language model and the proposed FineVQ on the constructed FineVD dataset in terms of the quality attribute prediction task.
For \textit{``yes-or-no''} question, the accuracy of random selection is 50\%.
It can be observed that the performance of some LMMs is lower than the random selection, which manifests that these LMMs have poor quality-related low-level inference ability.
Among the five specific dimensions, almost all LMMs have similar performance for the ``artifact'' dimension, which manifests that this dimension is difficult for all LMMs.
Moreover, different LMMs have different advantages, \textit{e.g.}, Qwen2-VL\cite{Qwen2VL} has good performance on the color distortion judgment, while having relatively poor ability to judge the blur distortion; LLaVA-NeXT~\cite{li2024llavanext-ablations} achieves the best performance in judging the color distortion, but its ability to judge the color distortion is relatively poor.
For \textit{``which''} question, the accuracy of random selection is 25\%.
Although some LMMs, such as VideoLLaMA2~\cite{damonlpsg2024videollama2} and MiniCPM-V~\cite{yao2024minicpm} have relatively good performance for \textit{``yes-or-no''} questions, we observe that almost all LMMs perform poorly in selecting the distortion type and judging the most influential distortion.
Finally, our FineVQ achieves the best performance for almost all tasks compared with state-of-the-art LMMs, which demonstrates the superiority of our model.

\begin{table}

\caption{The cross-dataset evaluation results. The models are trained on our FineVD and tested on other datasets.}
 \vspace{-2mm}
\resizebox{\linewidth}{!}{\begin{tabular}{lcccccc}
   \toprule
 \textit{Test on:} & \multicolumn{2}{c}{KoNViD-1k~\cite{hosu2017konstanz}} & \multicolumn{2}{c}{YouTtbe-UGC~\cite{wang2019youtube}}  & \multicolumn{2}{c}{LIVE-VQC~\cite{sinno2018large}}\\  \cmidrule(lr){2-3} \cmidrule(lr){4-5} \cmidrule(lr){6-7}
 \textit{Train on:} FineVD   & SRCC$\uparrow$&PLCC$\uparrow$ & SRCC$\uparrow$&PLCC$\uparrow$ & SRCC$\uparrow$&PLCC$\uparrow$\\  
     \midrule
   SimpleVQA~\cite{sun2022deep} &0.4818&0.3402& 0.7421&0.7231  & 0.5922&0.6137\\
   FastVQA~\cite{wu2022fast} &0.6793&0.6490 & 0.5607&0.5651  & 0.6767&0.6817     \\ 
  
\rowcolor{gray!20} FineVQ (Ours)   &\textbf{0.7171}&\textbf{0.7312} & \textbf{0.7797}&\textbf{0.7880}&\textbf{0.7417}&\textbf{0.7620}\\ \bottomrule
  \end{tabular}}
    \vspace{-1.5em}
  \label{tab4_cross2}
  \end{table}
\begin{table*}
\vspace{-0.3em}
\renewcommand\arraystretch{1.1}
  \caption{Ablation study of FineVQ on three datasets, including our FineVD, LIVE VQC \cite{sinno2018large}, and YouTube-UGC~\cite{wang2019youtube}.}

  \vspace{-0.7em}
  \resizebox{0.95\textwidth}{!}{
  \begin{tabular}{ccccc| ccc cccccc}
    \toprule
    \multicolumn{5}{c}{Strategy} & \multicolumn{3}{c}{FineVD} & \multicolumn{3}{c}{LIVE VQC \cite{sinno2018large}} & \multicolumn{3}{c}{YouTube-UGC \cite{wang2019youtube}} \\
    \cmidrule(r){1-5} \cmidrule(r){6-8} \cmidrule(r){9-11} \cmidrule(r){12-14}
    spatial  & motion   & LoRA$_{r=8}$ (vision) & LoRA$_{r=8}$ (llm) & LoRA$_{r=16}$     & SRCC     & PLCC     & KRCC          & SRCC       & PLCC    & KRCC          & SRCC     & PLCC   & KRCC      \\
    \midrule[0.3pt]
    \ding{52} & & &  &  &0.8210  & 0.8434 & 0.6356  &0.7832  & 0.8206 &  0.6065 &0.8084  & 0.8175 & 0.6149 \\
    \ding{52} & \ding{52} &  &  &  &0.8264 & 0.8459 & 0.6427 &0.8061  &0.8431  &  0.6294&0.8349  & 0.8525 & 0.6574\\
    \ding{52} & \ding{52} & \ding{52} &  &  & 0.8345  & 0.8478 & 0.6497  & 0.8207 & 0.8475 &0.6316 & 0.8442 & 0.8476   & 0.6600 \\
    \ding{52} & \ding{52} &  & \ding{52} &  & 0.8365 & 0.8549 & 0.6521 &0.8224  &0.8261  &0.6426& 0.8640 &0.8685  &  0.6842\\
    \ding{52} & \ding{52} & \ding{52} & \ding{52} &  & 0.8838 & 0.8874 & 0.7128 &0.8624  &0.8661  &0.6826  &0.8921  &0.9039  & 0.7212 \\
    \rowcolor{gray!20} \ding{52} & \ding{52} & \ding{52} & \ding{52} & \ding{52} &\textbf{0.8887} & \textbf{0.8964} & \textbf{0.7176}& \textbf{0.8951} & \textbf{0.8950} &  \textbf{0.7297}&  \textbf{0.9107}& \textbf{0.9141} & \textbf{0.7451} \\
    \bottomrule
  \end{tabular}\label{tab5_abl}
  }
  \vspace{-0.6em}
  \centering
\end{table*}
\begin{table*}
\setlength{\tabcolsep}{0.3em}
\renewcommand\arraystretch{1.2}
  \caption{Running time comparison with other methods.}
  \vspace{-0.7em}
  \resizebox{1\textwidth}{!}{
  \begin{tabular}{l | cccccccccccc}
    \toprule
    
    Model & NIQE~\cite{niqe} & QAC~\cite{qac} & HOSA~\cite{hosa} & TLVQM~\cite{korhonen2019two} & VIDEVAL~\cite{tu2021ugc} & RAPIQUE~\cite{tu2021rapique} & VSFA~\cite{li2019quality} & GSTVQA~\cite{chen2021learning} & SimpleVQA~\cite{sun2022deep} & FastVQA~\cite{wu2022fast} & Dover~\cite{wu2023exploring} & FineVQ \\
    \midrule[0.3pt]
    Time(s)$\downarrow$ & 918.4 & 369.0 & 250.0 & 848.2 & 1881 & 152.7 & 95.05 & 164.1 & 71.62 & \textbf{13.12} & \textbf{9.006} & \textbf{18.57} \\
    FPS$\uparrow$ & 2.613 & 6.505 & 9.600 & 2.830 & 1.277 & 15.72 & 25.26 & 14.63 & 33.52 & \textbf{182.9} & \textbf{266.5} & \textbf{129.3} \\
    
    \bottomrule
  \end{tabular}}\label{tab6_time}
\vspace{-1.3em}
\centering
\end{table*}

\subsection{Evaluation on Existing UGC-VQA Databases}

We further test the performance of the proposed FineVQ on other six VQA benchmark datasets, including LIVE-YT-Gaming~\cite{yu2023subjective}, KoNViD-1k~\cite{hosu2017konstanz}, YouTube-UGC~\cite{wang2019youtube}, LIVE-VQC~\cite{sinno2018large}, LSVQ\_test~\cite{ying2021patch}, LSVQ\_1080p~\cite{ying2021patch}.
We follow the settings used in previous studies to conduct the experiments \cite{sun2022deep,wu2022fast,wu2023exploring,lu2024kvq}.
Specifically, for LSVQ \cite{ying2021patch}, we follow the public training/test split setting to validate our method, which is trained on the training set of LSVQ and tested on the LSVQ$_\text{test}$ and LSVQ$_\text{1080p}$ test sets.
For the rest databases, we follow the previous principles, which split the dataset into training and test sets with a ratio of 4:1 and conduct multi-round cross-validation experiments.

Table \ref{tab3_sota} demonstrates the results of our proposed FineVQ and other state-of-the-art VQA methods on the six VQA benchmark datasets.
It can be observed that FineVQ achieves state-of-the-art performance on all six datasets in terms of both SRCC and PLCC metrics.
Specifically, the proposed FineVQ achieves 3.8\% performance improvement compared to KSVQE \cite{lu2024kvq} on LSVQ$_\text{1080p}$, demonstrating the effectiveness of FineVQ in high-resolution video quality assessment.

\subsection{Cross-dataset Evaluation}
We also conduct two cross-database evaluations, including (1) training on other datasets and testing on FineVD, and (2) training on FineVD and testing on other datasets. The results are shown in Table \ref{tab4_cross1} and Table \ref{tab4_cross2}, respectively. 

It can be concluded from the tables that, compared to the other two state-of-the-art VQA models, our proposed FineVQ model exhibits superior performance across both experimental settings, demonstrating the superior generalization capability of our model. Furthermore, comparing the generalization performance in Table \ref{tab4_cross1} and Table \ref{tab4_cross2}, we find that our FineVD database is more challenging, as models trained on FineVD typically perform better on other datasets but exhibit poorer results in the reverse setting. This validates the diversity and broad distribution of our dataset, which suggests that training on FineVD should result in models with enhanced generalization ability.

\subsection{Ablation Study}
We conduct ablation studies to validate the utility of the core components of FineVQ. The results are shown in Table \ref{tab5_abl}. The experiments are conducted on three datasets, including FineVD, LIVE-VQC \cite{sinno2018large} and YouTube-UGC \cite{wang2019youtube}, with consistent data partitioning and parameter settings.

\paragraph{Effectiveness of Motion Features.}
We first conduct ablation studies on the introduced motion feature extractor to validate the effectiveness of the extracted meticulous temporal features. Comparing the performance in the first two rows in Table \ref{tab5_abl}, introducing motion features yields a significant improvement over the baseline for all datasets, which highlights the importance of integrating motion features into the proposed model.

\paragraph{Effectiveness of LoRA Adaptation.}
We then demonstrate the effectiveness of LoRA adaptation, as shown in rows 3 to 5 in Table \ref{tab5_abl}. It can be observed that employing LoRA to either the image encoder (LoRA$_{r=8}$(vision)) or to the LLM (LoRA$_{r=8}$(llm)) enhances the performance of the proposed model. Moreover, combining both adaptations leads to a substantial improvement in performance.

\paragraph{Rank $r$ in LoRA.}
Finally, we compare the influence of different rank $r$ on the performance of FineVQ.
As shown in the last two rows in Table \ref{tab5_abl}, using $r=16$ generally performs better than using $r=8$ on all databases, which not only further validates the effect of LoRA, but also manifests the importance of task-specific adaptation.

\paragraph{Runtime.}
We also compare the runtime and FPS of state-of-the-art models on 10 2K 240-frame videos with an A6000 card or matlab and show results in Table \ref{tab6_time}. FineVQ achieves comparable results.

\section{Conclusion}
In this paper, we conduct a large-scale fine-grained UGC video quality assessment study.
Specifically, we first present FineVD, a large-scale VQA dataset containing 6104 UGC videos with fine-grained quality scores and descriptions from multiple dimensions.
Based on the database, we propose FineVQ to learn the fine-grained quality of UGC videos in a \textit{one-for-all} fashion via instruction tuning, which has excellent capabilities of quality rating, quality scoring, and quality attribution.
Extensive experiments on FineVD and other commonly used UGC-VQA datasets manifest that our FineVQ achieves state-of-the-art performance and can produce fine-grained video quality results.

\section*{Acknowledgment}
This work was supported in part by the NSFC 62401365, 62225112, 62271312, 62132006, 62271308, in part by the STCSM 24ZR1432000, 24511106902, 22511105700, 22DZ2229005, and in part by the 111 plan BP0719010.

{
    \small
    \bibliographystyle{ieeenat_fullname}
    \bibliography{main}

\begin{thebibliography}{81}
\providecommand{\natexlab}[1]{#1}
\providecommand{\url}[1]{\texttt{#1}}
\expandafter\ifx\csname urlstyle\endcsname\relax
  \providecommand{\doi}[1]{doi: #1}\else
  \providecommand{\doi}{doi: \begingroup \urlstyle{rm}\Url}\fi

\bibitem[Achiam et~al.(2023)Achiam, Adler, Agarwal, Ahmad, Akkaya, Aleman, Almeida, Altenschmidt, Altman, Anadkat, et~al.]{achiam2023gpt}
Josh Achiam, Steven Adler, Sandhini Agarwal, Lama Ahmad, Ilge Akkaya, Florencia~Leoni Aleman, Diogo Almeida, Janko Altenschmidt, Sam Altman, Shyamal Anadkat, et~al.
\newblock Gpt-4 technical report.
\newblock \emph{arXiv preprint arXiv:2303.08774}, 2023.

\bibitem[Chen et~al.(2021{\natexlab{a}})Chen, Zhu, Li, Lu, Fan, and Wang]{chen2021learning}
Baoliang Chen, Lingyu Zhu, Guo Li, Fangbo Lu, Hongfei Fan, and Shiqi Wang.
\newblock Learning generalized spatial-temporal deep feature representation for no-reference video quality assessment.
\newblock \emph{IEEE Transactions on Circuits and Systems for Video Technology (TCSVT)}, 32\penalty0 (4):\penalty0 1903--1916, 2021{\natexlab{a}}.

\bibitem[Chen et~al.(2021{\natexlab{b}})Chen, Li, Wu, Dong, and Shi]{chen2021contrastive}
Pengfei Chen, Leida Li, Jinjian Wu, Weisheng Dong, and Guangming Shi.
\newblock Contrastive self-supervised pre-training for video quality assessment.
\newblock \emph{IEEE Transactions on Image Processing (TIP)}, 31:\penalty0 458--471, 2021{\natexlab{b}}.

\bibitem[Chen et~al.(2021{\natexlab{c}})Chen, Li, Wu, Dong, and Shi]{chen2021unsupervised}
Pengfei Chen, Leida Li, Jinjian Wu, Weisheng Dong, and Guangming Shi.
\newblock Unsupervised curriculum domain adaptation for no-reference video quality assessment.
\newblock In \emph{Proceedings of the IEEE International Conference on Computer Vision (ICCV)}, pages 5178--5187, 2021{\natexlab{c}}.

\bibitem[Chen et~al.(2023)Chen, Liu, and Ding]{chen2023x}
Yixiong Chen, Li Liu, and Chris Ding.
\newblock X-iqe: explainable image quality evaluation for text-to-image generation with visual large language models.
\newblock \emph{arXiv preprint arXiv:2305.10843}, 2023.

\bibitem[Chen et~al.(2024)Chen, Wu, Wang, Su, Chen, Xing, Zhong, Zhang, Zhu, Lu, et~al.]{chen2024internvl}
Zhe Chen, Jiannan Wu, Wenhai Wang, Weijie Su, Guo Chen, Sen Xing, Muyan Zhong, Qinglong Zhang, Xizhou Zhu, Lewei Lu, et~al.
\newblock Internvl: Scaling up vision foundation models and aligning for generic visual-linguistic tasks.
\newblock In \emph{Proceedings of the IEEE Conference on Computer Vision and Pattern Recognition (CVPR)}, pages 24185--24198, 2024.

\bibitem[Cheng et~al.(2024)Cheng, Leng, Zhang, Xin, Li, Chen, Zhu, Zhang, Luo, Zhao, and Bing]{damonlpsg2024videollama2}
Zesen Cheng, Sicong Leng, Hang Zhang, Yifei Xin, Xin Li, Guanzheng Chen, Yongxin Zhu, Wenqi Zhang, Ziyang Luo, Deli Zhao, and Lidong Bing.
\newblock Videollama 2: Advancing spatial-temporal modeling and audio understanding in video-llms.
\newblock \emph{arXiv preprint arXiv:2406.07476}, 2024.

\bibitem[Dao()]{daoflashattention}
Tri Dao.
\newblock Flashattention-2: Faster attention with better parallelism and work partitioning.
\newblock In \emph{Proceedings of the International Conference on Learning Representations (ICLR)}.

\bibitem[Dao et~al.(2022)Dao, Fu, Ermon, Rudra, and R{\'e}]{dao2022flashattention}
Tri Dao, Dan Fu, Stefano Ermon, Atri Rudra, and Christopher R{\'e}.
\newblock Flashattention: Fast and memory-efficient exact attention with io-awareness.
\newblock \emph{Proceedings of the Advances in Neural Information Processing Systems (NeurIPS)}, 35:\penalty0 16344--16359, 2022.

\bibitem[De~Simone et~al.(2010)De~Simone, Tagliasacchi, Naccari, Tubaro, and Ebrahimi]{de2010h}
Francesca De~Simone, Marco Tagliasacchi, Matteo Naccari, Stefano Tubaro, and Touradj Ebrahimi.
\newblock A h. 264/avc video database for the evaluation of quality metrics.
\newblock In \emph{Proceedings of the IEEE International Conference on Acoustics, Speech and Signal Processing (ICASSP)}, pages 2430--2433, 2010.

\bibitem[Duan et~al.(2022{\natexlab{a}})Duan, Min, Zhu, Zhai, Yang, and Le~Callet]{duan2022confusing}
Huiyu Duan, Xiongkuo Min, Yucheng Zhu, Guangtao Zhai, Xiaokang Yang, and Patrick Le~Callet.
\newblock Confusing image quality assessment: Toward better augmented reality experience.
\newblock \emph{IEEE Transactions on Image Processing (TIP)}, 31:\penalty0 7206--7221, 2022{\natexlab{a}}.

\bibitem[Duan et~al.(2022{\natexlab{b}})Duan, Shen, Min, Tian, Jung, Yang, and Zhai]{duan2022develop}
Huiyu Duan, Wei Shen, Xiongkuo Min, Yuan Tian, Jae-Hyun Jung, Xiaokang Yang, and Guangtao Zhai.
\newblock Develop then rival: A human vision-inspired framework for superimposed image decomposition.
\newblock \emph{IEEE Transactions on Multimedia (TMM)}, 25:\penalty0 4267--4281, 2022{\natexlab{b}}.

\bibitem[Duan et~al.(2024)Duan, Min, Wu, Shen, and Zhai]{duan2024uniprocessor}
Huiyu Duan, Xiongkuo Min, Sijing Wu, Wei Shen, and Guangtao Zhai.
\newblock Uniprocessor: A text-induced unified low-level image processor.
\newblock In \emph{Proceedings of the European Conference on Computer Vision (ECCV)}, 2024.

\bibitem[Feichtenhofer et~al.(2019)Feichtenhofer, Fan, Malik, and He]{feichtenhofer2019slowfast}
Christoph Feichtenhofer, Haoqi Fan, Jitendra Malik, and Kaiming He.
\newblock Slowfast networks for video recognition.
\newblock In \emph{Proceedings of the IEEE International Conference on Computer Vision (ICCV)}, pages 6202--6211, 2019.

\bibitem[Ge et~al.(2024)Ge, Sun, Zhang, Li, Ji, Sun, Jui, Min, and Zhai]{ge2024lmm}
Qihang Ge, Wei Sun, Yu Zhang, Yunhao Li, Zhongpeng Ji, Fengyu Sun, Shangling Jui, Xiongkuo Min, and Guangtao Zhai.
\newblock Lmm-vqa: Advancing video quality assessment with large multimodal models.
\newblock \emph{arXiv preprint arXiv:2408.14008}, 2024.

\bibitem[Ghadiyaram et~al.(2017)Ghadiyaram, Pan, Bovik, Moorthy, Panda, and Yang]{ghadiyaram2017capture}
Deepti Ghadiyaram, Janice Pan, Alan~C Bovik, Anush~Krishna Moorthy, Prasanjit Panda, and Kai-Chieh Yang.
\newblock In-capture mobile video distortions: A study of subjective behavior and objective algorithms.
\newblock \emph{IEEE Transactions on Circuits and Systems for Video Technology (TCSVT)}, 28\penalty0 (9):\penalty0 2061--2077, 2017.

\bibitem[Haiqiang et~al.(2021)Haiqiang, Gary, Shan, and C.-C.~Jay]{icme21}
Wang Haiqiang, Li Gary, Liu Shan, and Kuo C.-C.~Jay.
\newblock Icme 2021 ugc-vqa challenge.
\newblock \url{http://ugcvqa.com/}, 2021.
\newblock [Online].

\bibitem[Hosu et~al.(2017)Hosu, Hahn, Jenadeleh, Lin, Men, Szir{\'a}nyi, Li, and Saupe]{hosu2017konstanz}
Vlad Hosu, Franz Hahn, Mohsen Jenadeleh, Hanhe Lin, Hui Men, Tam{\'a}s Szir{\'a}nyi, Shujun Li, and Dietmar Saupe.
\newblock The konstanz natural video database (konvid-1k).
\newblock In \emph{Proceedings of the IEEE International Conference on Quality of Multimedia Experience (QoMEX)}, pages 1--6, 2017.

\bibitem[Hu et~al.(2021)Hu, Wallis, Allen-Zhu, Li, Wang, Wang, Chen, et~al.]{hulora}
Edward~J Hu, Phillip Wallis, Zeyuan Allen-Zhu, Yuanzhi Li, Shean Wang, Lu Wang, Weizhu Chen, et~al.
\newblock Lora: Low-rank adaptation of large language models.
\newblock In \emph{Proceedings of the International Conference on Learning Representations (ICLR)}, 2021.

\bibitem[Inc.(2010)]{bilibili}
Bilibili Inc.
\newblock Bilibili.
\newblock \url{https://www.bilibili.com}, 2010.

\bibitem[Kancharla and Channappayya(2021)]{kancharla2021completely}
Parimala Kancharla and Sumohana~S Channappayya.
\newblock Completely blind quality assessment of user generated video content.
\newblock \emph{IEEE Transactions on Image Processing (TIP)}, 31:\penalty0 263--274, 2021.

\bibitem[Korhonen(2019)]{korhonen2019two}
Jari Korhonen.
\newblock Two-level approach for no-reference consumer video quality assessment.
\newblock \emph{IEEE Transactions on Image Processing (TIP)}, 28\penalty0 (12):\penalty0 5923--5938, 2019.

\bibitem[Li et~al.(2024{\natexlab{a}})Li, Zhang, Zhang, Guo, Zhang, Zhang, Li, Liu, and Li]{li2024llavanext-ablations}
Bo Li, Hao Zhang, Kaichen Zhang, Dong Guo, Yuanhan Zhang, Renrui Zhang, Feng Li, Ziwei Liu, and Chunyuan Li.
\newblock Llava-next: What else influences visual instruction tuning beyond data?, 2024{\natexlab{a}}.

\bibitem[Li et~al.(2019)Li, Jiang, and Jiang]{li2019quality}
Dingquan Li, Tingting Jiang, and Ming Jiang.
\newblock Quality assessment of in-the-wild videos.
\newblock In \emph{Proceedings of the ACM International Conference on Multimedia (ACM MM)}, pages 2351--2359, 2019.

\bibitem[Li et~al.(2022)Li, Li, Xiong, and Hoi]{li2022blip}
Junnan Li, Dongxu Li, Caiming Xiong, and Steven Hoi.
\newblock Blip: Bootstrapping language-image pre-training for unified vision-language understanding and generation.
\newblock In \emph{Proceedings of the International Conference on Machine Learning (ICML)}, pages 12888--12900. PMLR, 2022.

\bibitem[Li et~al.(2023)Li, Li, Savarese, and Hoi]{li2023blip}
Junnan Li, Dongxu Li, Silvio Savarese, and Steven Hoi.
\newblock Blip-2: Bootstrapping language-image pre-training with frozen image encoders and large language models.
\newblock In \emph{Proceedings of the International Conference on Machine Learning (ICML)}, pages 19730--19742. PMLR, 2023.

\bibitem[Li et~al.(2024{\natexlab{b}})Li, Wang, He, Li, Wang, Liu, Wang, Xu, Chen, Luo, et~al.]{li2024mvbench}
Kunchang Li, Yali Wang, Yinan He, Yizhuo Li, Yi Wang, Yi Liu, Zun Wang, Jilan Xu, Guo Chen, Ping Luo, et~al.
\newblock Mvbench: A comprehensive multi-modal video understanding benchmark.
\newblock In \emph{Proceedings of the IEEE Conference on Computer Vision and Pattern Recognition (CVPR)}, pages 22195--22206, 2024{\natexlab{b}}.

\bibitem[Li et~al.(2020)Li, Meng, Zhang, Wang, Wang, and Ma]{li2020ugc}
Yang Li, Shengbin Meng, Xinfeng Zhang, Shiqi Wang, Yue Wang, and Siwei Ma.
\newblock Ugc-video: Perceptual quality assessment of user-generated videos.
\newblock In \emph{Proceedings of the IEEE Conference on Multimedia Information Processing and Retrieval (MIPR)}, pages 35--38, 2020.

\bibitem[Liao et~al.(2022)Liao, Xu, Wu, Chen, Sun, Yan, and Lin]{liao2022exploring}
Liang Liao, Kangmin Xu, Haoning Wu, Chaofeng Chen, Wenxiu Sun, Qiong Yan, and Weisi Lin.
\newblock Exploring the effectiveness of video perceptual representation in blind video quality assessment.
\newblock In \emph{Proceedings of the ACM International Conference on Multimedia (ACM MM)}, pages 837--846, 2022.

\bibitem[Lin et~al.(2023)Lin, Zhu, Ye, Ning, Jin, and Yuan]{lin2023video}
Bin Lin, Bin Zhu, Yang Ye, Munan Ning, Peng Jin, and Li Yuan.
\newblock Video-llava: Learning united visual representation by alignment before projection.
\newblock \emph{arXiv preprint arXiv:2311.10122}, 2023.

\bibitem[Liu et~al.(2024{\natexlab{a}})Liu, Li, Li, and Lee]{liu2024improved}
Haotian Liu, Chunyuan Li, Yuheng Li, and Yong~Jae Lee.
\newblock Improved baselines with visual instruction tuning.
\newblock In \emph{Proceedings of the IEEE Conference on Computer Vision and Pattern Recognition (CVPR)}, pages 26296--26306, 2024{\natexlab{a}}.

\bibitem[Liu et~al.(2024{\natexlab{b}})Liu, Li, Wu, and Lee]{liu2024visual}
Haotian Liu, Chunyuan Li, Qingyang Wu, and Yong~Jae Lee.
\newblock Visual instruction tuning.
\newblock \emph{Proceedings of the Advances in Neural Information Processing Systems (NeurIPS)}, 36, 2024{\natexlab{b}}.

\bibitem[Liu et~al.(2021)Liu, Wu, Li, Dong, Zhang, and Shi]{liu2021spatiotemporal}
Yongxu Liu, Jinjian Wu, Leida Li, Weisheng Dong, Jinpeng Zhang, and Guangming Shi.
\newblock Spatiotemporal representation learning for blind video quality assessment.
\newblock \emph{IEEE Transactions on Circuits and Systems for Video Technology (TCSVT)}, 32\penalty0 (6):\penalty0 3500--3513, 2021.

\bibitem[Lu et~al.(2023)Lu, Li, Liu, Yang, Gao, and Shen]{lu2023empirical}
Yadong Lu, Chunyuan Li, Haotian Liu, Jianwei Yang, Jianfeng Gao, and Yelong Shen.
\newblock An empirical study of scaling instruct-tuned large multimodal models.
\newblock \emph{arXiv preprint arXiv:2309.09958}, 2023.

\bibitem[Lu et~al.(2024)Lu, Li, Pei, Yuan, Xie, Qu, Sun, Zhou, and Chen]{lu2024kvq}
Yiting Lu, Xin Li, Yajing Pei, Kun Yuan, Qizhi Xie, Yunpeng Qu, Ming Sun, Chao Zhou, and Zhibo Chen.
\newblock Kvq: Kwai video quality assessment for short-form videos.
\newblock In \emph{Proceedings of the IEEE Conference on Computer Vision and Pattern Recognition (CVPR)}, pages 25963--25973, 2024.

\bibitem[Maaz et~al.(2024)Maaz, Rasheed, Khan, and Khan]{Maaz2023VideoChatGPT}
Muhammad Maaz, Hanoona Rasheed, Salman Khan, and Fahad~Shahbaz Khan.
\newblock Video-chatgpt: Towards detailed video understanding via large vision and language models.
\newblock In \emph{Proceedings of the Annual Meeting of the Association for Computational Linguistics (ACL)}, 2024.

\bibitem[Min et~al.(2024)Min, Duan, Sun, Zhu, and Zhai]{min2024perceptual}
Xiongkuo Min, Huiyu Duan, Wei Sun, Yucheng Zhu, and Guangtao Zhai.
\newblock Perceptual video quality assessment: A survey.
\newblock \emph{Science China Information Sciences}, 67\penalty0 (11):\penalty0 211301, 2024.

\bibitem[Mitra and Soundararajan(2022)]{mitra2022multiview}
Shankhanil Mitra and Rajiv Soundararajan.
\newblock Multiview contrastive learning for completely blind video quality assessment of user generated content.
\newblock In \emph{Proceedings of the ACM International Conference on Multimedia (ACM MM)}, pages 1914--1924, 2022.

\bibitem[Mittal et~al.(2012)Mittal, Soundararajan, and Bovik]{niqe}
Anish Mittal, Rajiv Soundararajan, and Alan~C Bovik.
\newblock Making a “completely blind” image quality analyzer.
\newblock \emph{IEEE Signal Processing Letters (SPL)}, 20\penalty0 (3):\penalty0 209--212, 2012.

\bibitem[Mittal et~al.(2015)Mittal, Saad, and Bovik]{mittal2015completely}
Anish Mittal, Michele~A Saad, and Alan~C Bovik.
\newblock A completely blind video integrity oracle.
\newblock \emph{IEEE Transactions on Image Processing (TIP)}, 25\penalty0 (1):\penalty0 289--300, 2015.

\bibitem[Moorthy et~al.(2012)Moorthy, Choi, Bovik, and De~Veciana]{moorthy2012video}
Anush~Krishna Moorthy, Lark~Kwon Choi, Alan~Conrad Bovik, and Gustavo De~Veciana.
\newblock Video quality assessment on mobile devices: Subjective, behavioral and objective studies.
\newblock \emph{IEEE Journal of Selected Topics in Signal Processing (JSTSP)}, 6\penalty0 (6):\penalty0 652--671, 2012.

\bibitem[Nuutinen et~al.(2016)Nuutinen, Virtanen, Vaahteranoksa, Vuori, Oittinen, and H{\"a}kkinen]{nuutinen2016cvd2014}
Mikko Nuutinen, Toni Virtanen, Mikko Vaahteranoksa, Tero Vuori, Pirkko Oittinen, and Jukka H{\"a}kkinen.
\newblock Cvd2014—a database for evaluating no-reference video quality assessment algorithms.
\newblock \emph{IEEE Transactions on Image Processing (TIP)}, 25\penalty0 (7):\penalty0 3073--3086, 2016.

\bibitem[Radford et~al.(2021)Radford, Kim, Hallacy, Ramesh, Goh, Agarwal, Sastry, Askell, Mishkin, Clark, et~al.]{radford2021learning}
Alec Radford, Jong~Wook Kim, Chris Hallacy, Aditya Ramesh, Gabriel Goh, Sandhini Agarwal, Girish Sastry, Amanda Askell, Pamela Mishkin, Jack Clark, et~al.
\newblock Learning transferable visual models from natural language supervision.
\newblock In \emph{Proceedings of the International Conference on Machine Learning (ICML)}, pages 8748--8763. PMLR, 2021.

\bibitem[Saad et~al.(2014)Saad, Bovik, and Charrier]{saad2014blind}
Michele~A Saad, Alan~C Bovik, and Christophe Charrier.
\newblock Blind prediction of natural video quality.
\newblock \emph{IEEE Transactions on Image Processing (TIP)}, 23\penalty0 (3):\penalty0 1352--1365, 2014.

\bibitem[Schuhmann et~al.(2022)Schuhmann, Beaumont, Vencu, Gordon, Wightman, Cherti, Coombes, Katta, Mullis, Wortsman, et~al.]{schuhmann2022laion}
Christoph Schuhmann, Romain Beaumont, Richard Vencu, Cade Gordon, Ross Wightman, Mehdi Cherti, Theo Coombes, Aarush Katta, Clayton Mullis, Mitchell Wortsman, et~al.
\newblock Laion-5b: An open large-scale dataset for training next generation image-text models.
\newblock \emph{Proceedings of the Advances in Neural Information Processing Systems (NeurIPS)}, 35:\penalty0 25278--25294, 2022.

\bibitem[Series(2012)]{series2012methodology}
BT Series.
\newblock Methodology for the subjective assessment of the quality of television pictures.
\newblock \emph{Recommendation ITU-R BT}, pages 500--13, 2012.

\bibitem[Seshadrinathan et~al.(2010)Seshadrinathan, Soundararajan, Bovik, and Cormack]{seshadrinathan2010study}
Kalpana Seshadrinathan, Rajiv Soundararajan, Alan~Conrad Bovik, and Lawrence~K Cormack.
\newblock Study of subjective and objective quality assessment of video.
\newblock \emph{IEEE Transactions on Image Processing (TIP)}, 19\penalty0 (6):\penalty0 1427--1441, 2010.

\bibitem[Sinno and Bovik(2018)]{sinno2018large}
Zeina Sinno and Alan~Conrad Bovik.
\newblock Large-scale study of perceptual video quality.
\newblock \emph{IEEE Transactions on Image Processing (TIP)}, 28\penalty0 (2):\penalty0 612--627, 2018.

\bibitem[Sun et~al.(2022)Sun, Min, Lu, and Zhai]{sun2022deep}
Wei Sun, Xiongkuo Min, Wei Lu, and Guangtao Zhai.
\newblock A deep learning based no-reference quality assessment model for ugc videos.
\newblock In \emph{Proceedings of the ACM International Conference on Multimedia (ACM MM)}, pages 856--865, 2022.

\bibitem[Sun et~al.(2024)Sun, Wen, Min, Lan, Zhai, and Ma]{sun2024analysis}
Wei Sun, Wen Wen, Xiongkuo Min, Long Lan, Guangtao Zhai, and Kede Ma.
\newblock Analysis of video quality datasets via design of minimalistic video quality models.
\newblock \emph{IEEE Transactions on Pattern Analysis and Machine Intelligence (TPAMI)}, 2024.

\bibitem[Team(2023)]{team2023internlm}
InternLM Team.
\newblock Internlm: A multilingual language model with progressively enhanced capabilities, 2023.

\bibitem[Touvron et~al.(2023)Touvron, Lavril, Izacard, Martinet, Lachaux, Lacroix, Rozi{\`e}re, Goyal, Hambro, Azhar, et~al.]{touvron2023llama}
Hugo Touvron, Thibaut Lavril, Gautier Izacard, Xavier Martinet, Marie-Anne Lachaux, Timoth{\'e}e Lacroix, Baptiste Rozi{\`e}re, Naman Goyal, Eric Hambro, Faisal Azhar, et~al.
\newblock Llama: Open and efficient foundation language models.
\newblock \emph{arXiv preprint arXiv:2302.13971}, 2023.

\bibitem[Tu et~al.(2021{\natexlab{a}})Tu, Wang, Birkbeck, Adsumilli, and Bovik]{tu2021ugc}
Zhengzhong Tu, Yilin Wang, Neil Birkbeck, Balu Adsumilli, and Alan~C Bovik.
\newblock Ugc-vqa: Benchmarking blind video quality assessment for user generated content.
\newblock \emph{IEEE Transactions on Image Processing (TIP)}, 30:\penalty0 4449--4464, 2021{\natexlab{a}}.

\bibitem[Tu et~al.(2021{\natexlab{b}})Tu, Yu, Wang, Birkbeck, Adsumilli, and Bovik]{tu2021rapique}
Zhengzhong Tu, Xiangxu Yu, Yilin Wang, Neil Birkbeck, Balu Adsumilli, and Alan~C Bovik.
\newblock Rapique: Rapid and accurate video quality prediction of user generated content.
\newblock \emph{IEEE Open Journal of Signal Processing}, 2:\penalty0 425--440, 2021{\natexlab{b}}.

\bibitem[Wang et~al.(2016)Wang, Gan, Hu, Lin, Jin, Song, Wang, Katsavounidis, Aaron, and Kuo]{wang2016mcl}
Haiqiang Wang, Weihao Gan, Sudeng Hu, Joe~Yuchieh Lin, Lina Jin, Longguang Song, Ping Wang, Ioannis Katsavounidis, Anne Aaron, and C-C~Jay Kuo.
\newblock Mcl-jcv: a jnd-based h. 264/avc video quality assessment dataset.
\newblock In \emph{Proceedings of the IEEE International Conference on Image Processing (ICIP)}, pages 1509--1513, 2016.

\bibitem[Wang et~al.(2024)Wang, Bai, Tan, Wang, Fan, Bai, Chen, Liu, Wang, Ge, Fan, Dang, Du, Ren, Men, Liu, Zhou, Zhou, and Lin]{Qwen2VL}
Peng Wang, Shuai Bai, Sinan Tan, Shijie Wang, Zhihao Fan, Jinze Bai, Keqin Chen, Xuejing Liu, Jialin Wang, Wenbin Ge, Yang Fan, Kai Dang, Mengfei Du, Xuancheng Ren, Rui Men, Dayiheng Liu, Chang Zhou, Jingren Zhou, and Junyang Lin.
\newblock Qwen2-vl: Enhancing vision-language model's perception of the world at any resolution.
\newblock \emph{arXiv preprint arXiv:2409.12191}, 2024.

\bibitem[Wang et~al.(2019)Wang, Inguva, and Adsumilli]{wang2019youtube}
Yilin Wang, Sasi Inguva, and Balu Adsumilli.
\newblock Youtube ugc dataset for video compression research.
\newblock In \emph{Proceedings of the IEEE International Workshop on Multimedia Signal Processing (MMSP)}, pages 1--5, 2019.

\bibitem[Wang et~al.(2021)Wang, Ke, Talebi, Yim, Birkbeck, Adsumilli, Milanfar, and Yang]{wang2021rich}
Yilin Wang, Junjie Ke, Hossein Talebi, Joong~Gon Yim, Neil Birkbeck, Balu Adsumilli, Peyman Milanfar, and Feng Yang.
\newblock Rich features for perceptual quality assessment of ugc videos.
\newblock In \emph{Proceedings of the IEEE Conference on Computer Vision and Pattern Recognition (CVPR)}, pages 13435--13444, 2021.

\bibitem[Wei et~al.(2022)Wei, Wang, Schuurmans, Bosma, Xia, Chi, Le, Zhou, et~al.]{wei2022chain}
Jason Wei, Xuezhi Wang, Dale Schuurmans, Maarten Bosma, Fei Xia, Ed Chi, Quoc~V Le, Denny Zhou, et~al.
\newblock Chain-of-thought prompting elicits reasoning in large language models.
\newblock \emph{Advances in neural information processing systems}, 35:\penalty0 24824--24837, 2022.

\bibitem[Wu et~al.(2022)Wu, Chen, Hou, Liao, Wang, Sun, Yan, and Lin]{wu2022fast}
Haoning Wu, Chaofeng Chen, Jingwen Hou, Liang Liao, Annan Wang, Wenxiu Sun, Qiong Yan, and Weisi Lin.
\newblock Fast-vqa: Efficient end-to-end video quality assessment with fragment sampling.
\newblock In \emph{Proceedings of the European Conference on Computer Vision (ECCV)}, pages 538--554. Springer, 2022.

\bibitem[Wu et~al.(2023{\natexlab{a}})Wu, Chen, Liao, Hou, Sun, Yan, Gu, and Lin]{wu2023neighbourhood}
Haoning Wu, Chaofeng Chen, Liang Liao, Jingwen Hou, Wenxiu Sun, Qiong Yan, Jinwei Gu, and Weisi Lin.
\newblock Neighbourhood representative sampling for efficient end-to-end video quality assessment.
\newblock \emph{IEEE Transactions on Pattern Analysis and Machine Intelligence (TPAMI)}, 2023{\natexlab{a}}.

\bibitem[Wu et~al.(2023{\natexlab{b}})Wu, Zhang, Liao, Chen, Hou, Wang, Sun, Yan, and Lin]{wu2023exploring}
Haoning Wu, Erli Zhang, Liang Liao, Chaofeng Chen, Jingwen Hou, Annan Wang, Wenxiu Sun, Qiong Yan, and Weisi Lin.
\newblock Exploring video quality assessment on user generated contents from aesthetic and technical perspectives.
\newblock In \emph{Proceedings of the IEEE International Conference on Computer Vision (ICCV)}, pages 20144--20154, 2023{\natexlab{b}}.

\bibitem[Wu et~al.(2023{\natexlab{c}})Wu, Zhang, Liao, Chen, Hou, Wang, Sun, Yan, and Lin]{wu2023towards}
Haoning Wu, Erli Zhang, Liang Liao, Chaofeng Chen, Jingwen Hou, Annan Wang, Wenxiu Sun, Qiong Yan, and Weisi Lin.
\newblock Towards explainable in-the-wild video quality assessment: a database and a language-prompted approach.
\newblock In \emph{Proceedings of the ACM International Conference on Multimedia (ACM MM)}, pages 1045--1054, 2023{\natexlab{c}}.

\bibitem[Wu et~al.(2024{\natexlab{a}})Wu, Zhang, Zhang, Chen, Liao, Wang, Li, Sun, Yan, Zhai, et~al.]{QBench}
Haoning Wu, Zicheng Zhang, Erli Zhang, Chaofeng Chen, Liang Liao, Annan Wang, Chunyi Li, Wenxiu Sun, Qiong Yan, Guangtao Zhai, et~al.
\newblock Q-bench: A benchmark for general-purpose foundation models on low-level vision.
\newblock In \emph{Proceedings of the International Conference on Learning Representations (ICLR)}, 2024{\natexlab{a}}.

\bibitem[Wu et~al.(2024{\natexlab{b}})Wu, Zhang, Zhang, Chen, Liao, Wang, Xu, Li, Hou, Zhai, et~al.]{Qinstruct}
Haoning Wu, Zicheng Zhang, Erli Zhang, Chaofeng Chen, Liang Liao, Annan Wang, Kaixin Xu, Chunyi Li, Jingwen Hou, Guangtao Zhai, et~al.
\newblock Q-instruct: Improving low-level visual abilities for multi-modality foundation models.
\newblock In \emph{Proceedings of the IEEE Conference on Computer Vision and Pattern Recognition (CVPR)}, pages 25490--25500, 2024{\natexlab{b}}.

\bibitem[Wu et~al.(2024{\natexlab{c}})Wu, Zhang, Zhang, Chen, Liao, Li, Gao, Wang, Zhang, Sun, et~al.]{QAlign}
Haoning Wu, Zicheng Zhang, Weixia Zhang, Chaofeng Chen, Liang Liao, Chunyi Li, Yixuan Gao, Annan Wang, Erli Zhang, Wenxiu Sun, et~al.
\newblock Q-align: Teaching lmms for visual scoring via discrete text-defined levels.
\newblock In \emph{Proceedings of the International Conference on Machine Learning (ICML)}, 2024{\natexlab{c}}.

\bibitem[Wu et~al.(2021)Wu, Liu, Li, Dong, and Shi]{wu2021no}
Jinjian Wu, Yongxu Liu, Leida Li, Weisheng Dong, and Guangming Shi.
\newblock No-reference video quality assessment with heterogeneous knowledge ensemble.
\newblock In \emph{Proceedings of the ACM International Conference on Multimedia (ACM MM)}, pages 4174--4182, 2021.

\bibitem[Xu et~al.(2016)Xu, Ye, Li, Du, Liu, and Doermann]{hosa}
Jingtao Xu, Peng Ye, Qiaohong Li, Haiqing Du, Yong Liu, and David Doermann.
\newblock Blind image quality assessment based on high order statistics aggregation.
\newblock \emph{IEEE Transactions on Image Processing (TIP)}, 25\penalty0 (9):\penalty0 4444--4457, 2016.

\bibitem[Xue et~al.(2013)Xue, Zhang, and Mou]{qac}
Wufeng Xue, Lei Zhang, and Xuanqin Mou.
\newblock Learning without human scores for blind image quality assessment.
\newblock In \emph{Proceedings of the IEEE Conference on Computer Vision and Pattern Recognition (CVPR)}, pages 995--1002, 2013.

\bibitem[Yao et~al.(2024)Yao, Yu, Zhang, Wang, Cui, Zhu, Cai, Li, Zhao, He, et~al.]{yao2024minicpm}
Yuan Yao, Tianyu Yu, Ao Zhang, Chongyi Wang, Junbo Cui, Hongji Zhu, Tianchi Cai, Haoyu Li, Weilin Zhao, Zhihui He, et~al.
\newblock Minicpm-v: A gpt-4v level mllm on your phone.
\newblock \emph{arXiv preprint arXiv:2408.01800}, 2024.

\bibitem[Ye et~al.(2024)Ye, Xu, Ye, Yan, Hu, Liu, Qian, Zhang, and Huang]{ye2024mplug}
Qinghao Ye, Haiyang Xu, Jiabo Ye, Ming Yan, Anwen Hu, Haowei Liu, Qi Qian, Ji Zhang, and Fei Huang.
\newblock mplug-owl2: Revolutionizing multi-modal large language model with modality collaboration.
\newblock In \emph{Proceedings of the IEEE Conference on Computer Vision and Pattern Recognition (CVPR)}, pages 13040--13051, 2024.

\bibitem[Ying et~al.(2021)Ying, Mandal, Ghadiyaram, and Bovik]{ying2021patch}
Zhenqiang Ying, Maniratnam Mandal, Deepti Ghadiyaram, and Alan Bovik.
\newblock Patch-vq:'patching up'the video quality problem.
\newblock In \emph{Proceedings of the IEEE Conference on Computer Vision and Pattern Recognition (CVPR)}, pages 14019--14029, 2021.

\bibitem[You(2021)]{you2021long}
Junyong You.
\newblock Long short-term convolutional transformer for no-reference video quality assessment.
\newblock In \emph{Proceedings of the ACM International Conference on Multimedia (ACM MM)}, pages 2112--2120, 2021.

\bibitem[You et~al.(2023)You, Li, Gu, Yin, Xue, and Dong]{you2023depicting}
Zhiyuan You, Zheyuan Li, Jinjin Gu, Zhenfei Yin, Tianfan Xue, and Chao Dong.
\newblock Depicting beyond scores: Advancing image quality assessment through multi-modal language models.
\newblock \emph{arXiv preprint arXiv:2312.08962}, 2023.

\bibitem[You et~al.(2024)You, Gu, Li, Cai, Zhu, Dong, and Xue]{you2024descriptive}
Zhiyuan You, Jinjin Gu, Zheyuan Li, Xin Cai, Kaiwen Zhu, Chao Dong, and Tianfan Xue.
\newblock Descriptive image quality assessment in the wild.
\newblock \emph{arXiv preprint arXiv:2405.18842}, 2024.

\bibitem[Yu et~al.(2021)Yu, Birkbeck, Wang, Bampis, Adsumilli, and Bovik]{yu2021predicting}
Xiangxu Yu, Neil Birkbeck, Yilin Wang, Christos~G Bampis, Balu Adsumilli, and Alan~C Bovik.
\newblock Predicting the quality of compressed videos with pre-existing distortions.
\newblock \emph{IEEE Transactions on Image Processing (TIP)}, 30:\penalty0 7511--7526, 2021.

\bibitem[Yu et~al.(2023)Yu, Ying, Birkbeck, Wang, Adsumilli, and Bovik]{yu2023subjective}
Xiangxu Yu, Zhenqiang Ying, Neil Birkbeck, Yilin Wang, Balu Adsumilli, and Alan~C Bovik.
\newblock Subjective and objective analysis of streamed gaming videos.
\newblock \emph{IEEE Transactions on Games}, 2023.

\bibitem[Yuan et~al.(2023)Yuan, Kong, Zheng, Sun, and Wen]{yuan2023capturing}
Kun Yuan, Zishang Kong, Chuanchuan Zheng, Ming Sun, and Xing Wen.
\newblock Capturing co-existing distortions in user-generated content for no-reference video quality assessment.
\newblock In \emph{Proceedings of the ACM International Conference on Multimedia (ACM MM)}, pages 1098--1107, 2023.

\bibitem[Zhang et~al.(2023)Zhang, Wu, Sun, Tu, Lu, Min, Chen, and Zhai]{zhang2023md}
Zicheng Zhang, Wei Wu, Wei Sun, Danyang Tu, Wei Lu, Xiongkuo Min, Ying Chen, and Guangtao Zhai.
\newblock Md-vqa: Multi-dimensional quality assessment for ugc live videos.
\newblock In \emph{Proceedings of the IEEE Conference on Computer Vision and Pattern Recognition (CVPR)}, pages 1746--1755, 2023.

\bibitem[Zheng et~al.(2022)Zheng, Tu, Zeng, Bovik, and Fan]{zheng2022completely}
Qi Zheng, Zhengzhong Tu, Xiaoyang Zeng, Alan~C Bovik, and Yibo Fan.
\newblock A completely blind video quality evaluator.
\newblock \emph{IEEE Signal Processing Letters (SPL)}, 29:\penalty0 2228--2232, 2022.

\bibitem[Zhu et~al.(2024)Zhu, Chen, Shen, Li, and Elhoseiny]{zhuminigpt}
Deyao Zhu, Jun Chen, Xiaoqian Shen, Xiang Li, and Mohamed Elhoseiny.
\newblock Minigpt-4: Enhancing vision-language understanding with advanced large language models.
\newblock In \emph{Proceedings of the International Conference on Learning Representations (ICLR)}, 2024.

\end{thebibliography}
}

\appendix

\clearpage
\newpage

\twocolumn[{
\centering
\includegraphics[width=1\linewidth]{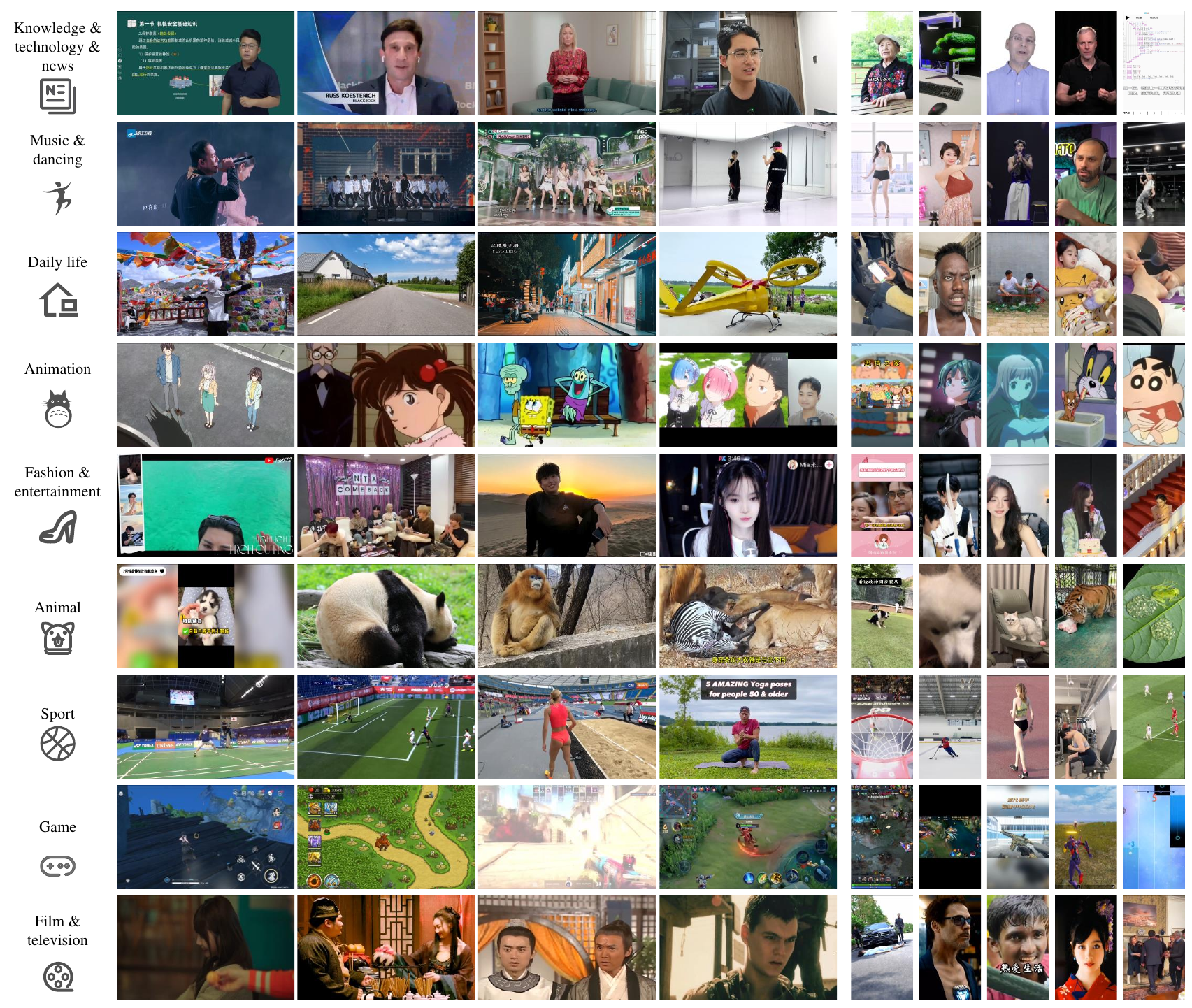}
\vspace{-0.6em}
\captionof{figure}{More examples of the on-demand UGC videos in our FineVD. Some videos are resized for better illustration.
}
\label{fig_supp:1}
\vspace{1.5em}
}]

\section{More Details of the Collected Videos}

\subsection{More Examples of the Collected Videos}
We first show more sample videos to illustrate the abundant content in the FineVD database.
As shown in Figure \ref{fig_supp:1}, for on-demand UGC videos, the content in FineVD covers \textit{knowledge \& technology \& news, music \& dancing, daily life, animation, fashion \& entertainment, animal, sport, game, film \& television} videos.
Moreover, all categories contain both traditional UGC videos (landscape videos) and short-form videos (portrait videos), which indicates the wide coverage of videos.
In particular, our FineVD also contains animation and game videos, which are typically ignored by previous databases \cite{sinno2018large,hosu2017konstanz,wang2019youtube}.

\begin{figure*}[t]\centering
\includegraphics[width=1\linewidth]{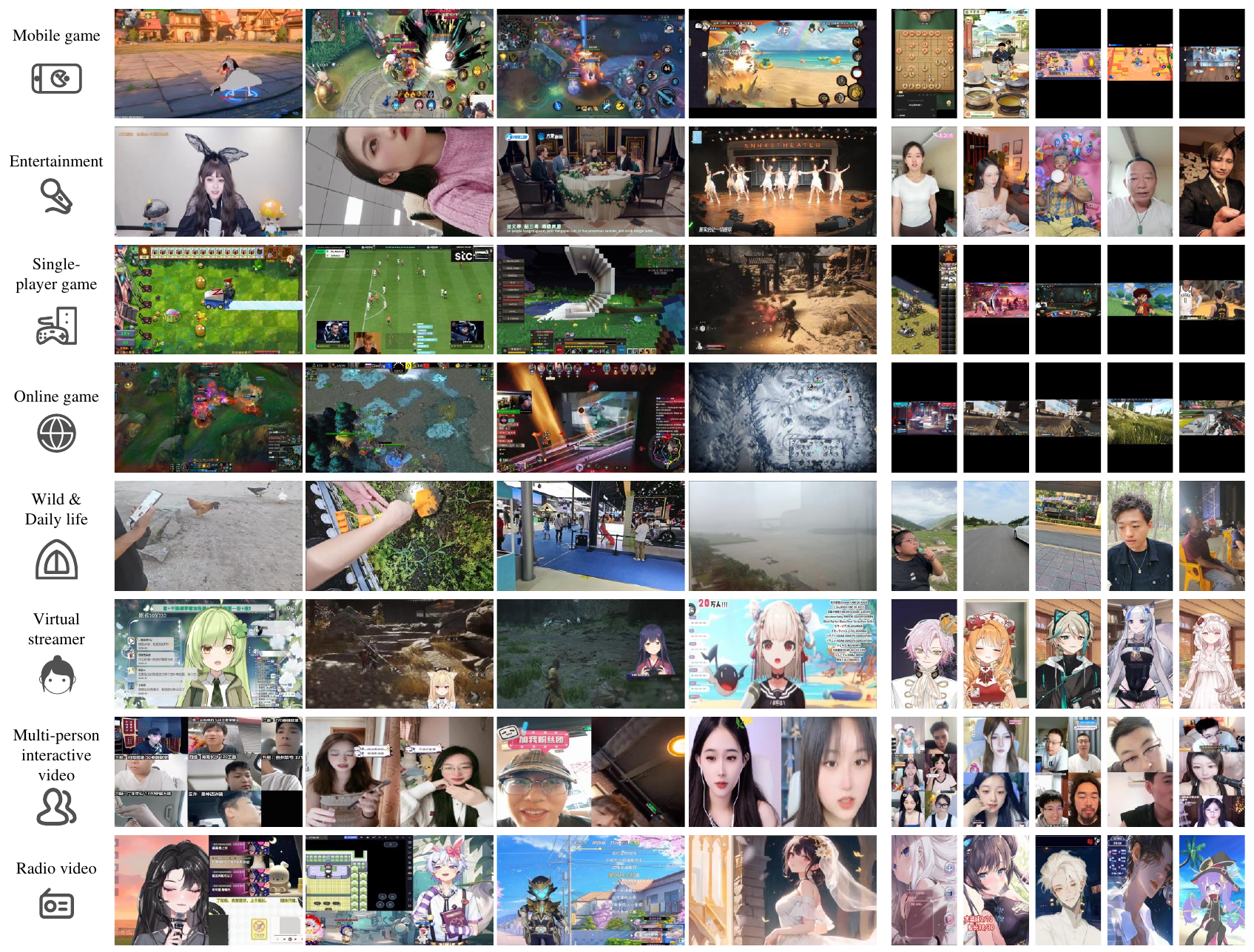}
\vspace{-0.6em}
\caption{More examples of the live-streaming UGC videos in our FineVD. Some videos are resized for better illustration.
}
\label{fig_supp:2}
\end{figure*}

Moreover, our FineVD also contains abundant live-streaming videos, which cover the categories of \textit{mobile game, entertainment, single-player game, online game, wild \& daily life, virtual streamer, multi-person interactive video, radio video}, as shown in Figure \ref{fig_supp:2}.
The live-streaming videos also contain both traditional-form videos and short-form videos.
In particular, the virtual streamer videos and the radio videos should be distinguished, since most virtual streamer videos contain moving virtual characters, while most radio videos are static characters or wallpapers.

We also compare the constructed FineVD database with other popular public UGC VQA datasets \cite{min2024perceptual}.
As shown in Table \ref{tab:supp_overview}, it can be observed that our FineVD database is the first UGC VQA database that contains multi-dimensional MOS annotations and fine-grained descriptions.
Moreover, compared with the recent two databases TaoLive \cite{zhang2023md} and KVQ \cite{lu2024kvq}, the FineVD is also established in a lab environment, but contains more diverse video content and more fine-grained annotations.

\begin{table*}[t]
    \centering
    \caption{An overview of popular public UGC VQA datasets.}
    \label{tab:supp_overview}
    \setlength{\tabcolsep}{0.35em}
    \resizebox{1\linewidth}{!}{
    \renewcommand{\arraystretch}{1.5}
    \begin{tabular}{l l l l l l l l l l l l l l}
    \toprule
        Database & Type & Year & \#Cont. & \#Total & Resolution & FR & Dur. & Format & Distortions & \#Subj. & \#Ratings & Data & Env. \\
        \hline
        
        CVD2014~\cite{nuutinen2016cvd2014} & Aut. & 2014 & 5 & 234 & 720p, 480p & 9-30 & 10-25 & AVI & In-capture & 210 & 30 & MOS & In-lab \\
        LIVE-Qualcomm~\cite{ghadiyaram2017capture} & Aut. & 2016 & 54 & 208 & 1080p & 30 & 15 & YUV & In-capture & 39 & 39 & MOS & In-lab \\
        \hline
        UGC-VIDEO~\cite{li2020ugc} & Syn.$+$Aut. & 2019 & 50 & 550 & 720p & 30 & 10 & N/A & UGC$+$compression & 30 & 30 & DMOS & In-lab \\
        LIVE-WC~\cite{yu2021predicting} & Syn.$+$Aut. & 2020 & 55 & 275 & 1080p & 30 & 10 & MP4 & UGC$+$compression & 40 & 40 & MOS & In-lab \\
        YT-UGC$+$(Subset)~\cite{wang2021rich} & Syn.$+$Aut. & 2021 & 189 & 567 & 1080p, 720p & Diverse & 20 & RAW$+$264 & UGC$+$compression & N/A & 30 & DMOS & In-lab \\
        ICME2021~\cite{icme21} & Syn.$+$Aut. & 2021 & 1000 & 8000 & N/A & N/A & N/A & N/A & UGC$+$compression & N/A & N/A & MOS & In-lab \\
        TaoLive~\cite{zhang2023md} & Syn.$+$Aut. & 2023 & 418 & 3762 & 1080p, 720p & 20 & 8 & MP4 & UGC$+$compression & 44 & 44 & MOS & In-lab \\
        KVQ~\cite{lu2024kvq} & Syn.$+$Aut. & 2024 & 600 & 3600 & Diverse & Diverse & 8 & MP4 & UGC$+$compression & 15 & 15 & MOS+Rank & In-lab \\
        \hline
        KoNViD-1k~\cite{hosu2017konstanz} & Aut. & 2017 & 1200 & 1200 & 540p & 24-30 & 8 & MP4 & In-the-wild & 642 & 114 & MOS$+$$\sigma$ & Crowd \\
        LIVE-VQC~\cite{sinno2018large} & Aut. & 2018 & 585 & 585 & 1080p-240p & 19-30 & 10 & MP4 & In-the-wild & 4776 & 240 & MOS & Crowd \\
        YouTube-UGC~\cite{wang2019youtube} & Aut. & 2019 & 1380 & 1380 & 4k-360p & 15-60 & 20 & MKV & In-the-wild & $>$8k & 123 & MOS$+$$\sigma$ & Crowd \\
        LSVQ~\cite{ying2021patch} & Aut. & 2021 &  39075 & 39075 & Diverse & Diverse & 5-12 & MP4 & In-the-wild & 6284 & 35 & MOS & Crowd \\
        \rowcolor{gray!20} FineVD (Ours) & Aut. & 2024 &  6104 & 6104 & Diverse & Diverse & 8 & MP4 & In-the-wild & 22 & 22 & MOS$\times$6$+$$\sigma$$\times$6$+$Descriptions & In-lab \\
    \bottomrule
    \end{tabular}
    }
    \raggedright
    \scalebox{0.69}{
    \begin{tabular}{l}
        Note: \#Cont.: The number of unique video contents. \quad \#Total: Total number of test video sequences. \quad FR: Framerate (in fps). \quad Dur.: Video duration/length (in seconds). \\
         \quad  \quad  \quad \#Subj.: Total number of subjects in the study. \quad \#Ratings: Average number of subjective ratings per video. \quad Env.: Subjective experiment environment. \\
         \quad  \quad  \quad In-lab: Experiment was conducted in a laboratory. \quad Crowd: Experiment was conducted by crowdsourcing. \quad Syn.: Synthetic. \quad Aut.: Authentic.
    \end{tabular}
    }
\end{table*}

\begin{figure}[t]\centering
\includegraphics[width=1\linewidth]{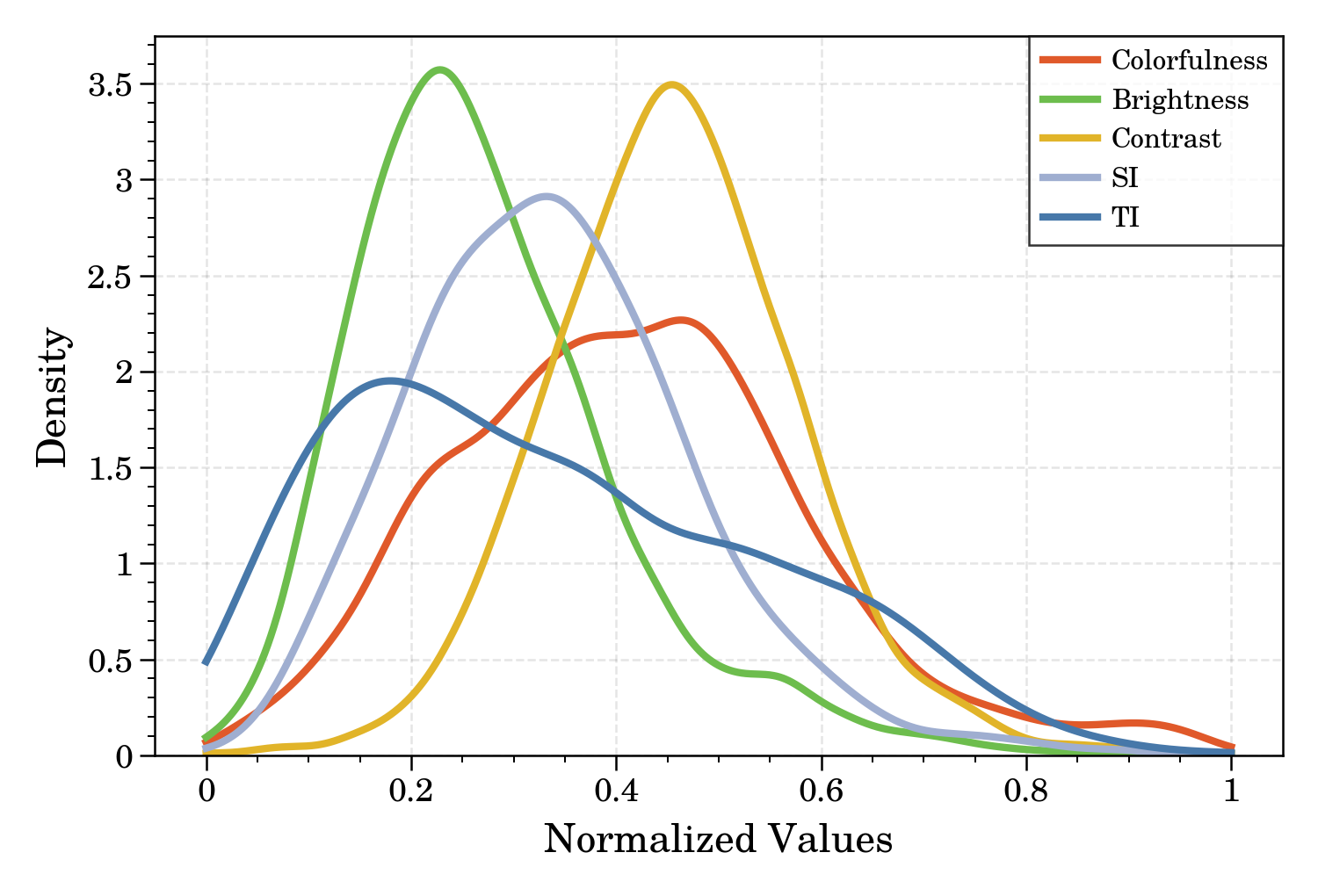}
\vspace{-0.6em}
\caption{The feature distribution of FineVD. SI and TI indicate spatial information and temporal information, respectively.
}
\label{fig_supp:3}
\end{figure}

\subsection{Feature Analysis}
As shown in Figure \ref{fig_supp:3}, the constructed FineVD database exhibits broad feature characteristics across five video quality-related features, including colorfulness, brightness, contrast, SI, TI.
It can be observed that the majority of features span a wide range of normalized values, indicating the feature diversity inherent in out database.
Specifically, the colorfulness coverage in Figure \ref{fig_supp:3} is relatively uniform, which further manifests the video content in FineVD is abundant.
Moreover, the SI and TI also cover a wide range, indicating that the video content in FineVD has both spatial richness and temporal richness.

\begin{figure*}[t]\centering
\includegraphics[width=1\linewidth]{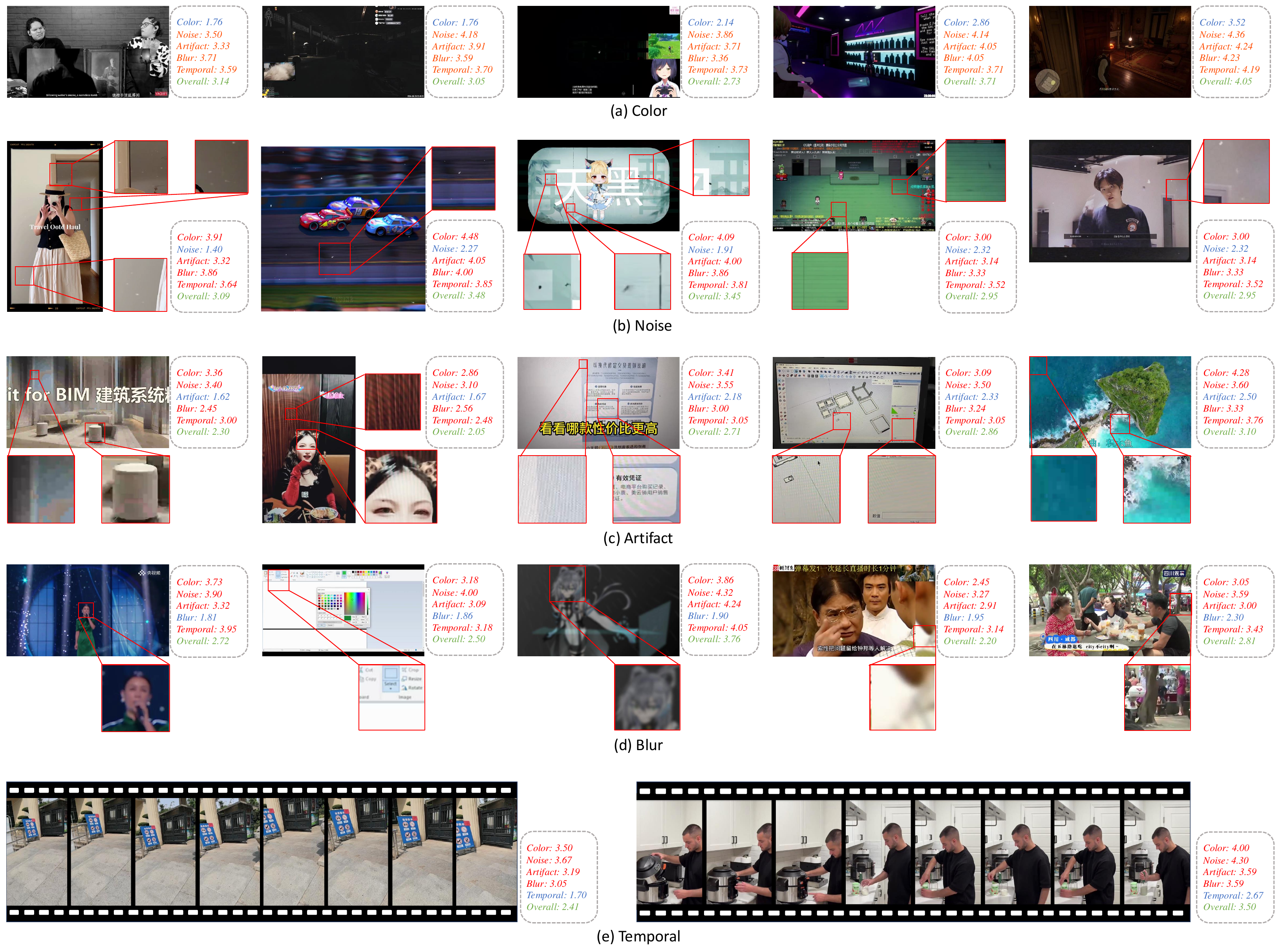}
\vspace{-0.6em}
\caption{Illustration of the differences between different dimensions.
}
\label{fig_supp:4}
\end{figure*}

\subsection{Distortion Analysis}

Figure \ref{fig_supp:4} illustrates the differences between different dimensions.
It can be observed that the five dimensions, \textit{i.e.,} color, noise, artifact, blur, and temporal are significantly distinct in some cases.
For example, as shown in the first row, the MOSs of the color dimension for the five videos are low, but the MOSs of other dimensions are relatively high.
Moreover, we also notice that the overall score is generally in the middle of the worst dimension and other dimensions, which also manifests that the overall quality is significantly affected by the most severe distortion.

\section{More Details of Subjective Study}

\subsection{Subjective Experiment Setup}

The subjective experiment is conducted among 22 subjects.
We use the 5-level rating method following the recommendation of ITU \cite{series2012methodology} to conduct the experiment, and the quality bar is labeled with five Likert adjectives, including ``Bad, Poor, Fair, Good and Excellent'', respectively.
Given the peculiarities of the fine-grained quality assessment, we advise the subjects to give their opinion scores following the instructions:

\paragraph{Noise dimension:}
(1) Severe: There are serious image noise or particles in the video. The noise is very obvious, resulting in unclear content and seriously affecting the viewing experience.
(2) Strong: The image noise in the video is obvious, affecting the viewing experience, but it does not affect the comprehensibility of the content.
(3) Mild: The noise in the video exists, but it is not too significant and may affect the details.
(4) Slight: In most cases, the image in the video has no obvious noise. The noise may exist for a short time.
(5) Undistorted: There is almost no visible noise problem in the video.

\paragraph{Artifact dimension:}
(1) Severe: There are serious artifacts in the video, obvious block distortion, compression distortion or other obvious visual defects, which significantly interfere with the viewing experience.
(2) Strong: The artifacts, flaws or distortions in the video are obvious, but they do not seriously affect the comprehensibility of the content.
(3) Mild: The artifacts, flaws or distortions in the video exist, but they are not too obvious and have a slight impact on the video content.
(4) Slight: The artifacts, flaws or distortions in the video are few and basically do not interfere with viewing. They may only be noticed in a few cases.
(5) Undistorted: The video has almost no visible artifacts, flaws or distortions.

\paragraph{Blur dimension:}
(1) Severe: The video is extremely blurry, details are difficult to discern, and even the main objects or people cannot be identified, showing obvious pixelation or blurring effects, which seriously affects the viewing experience.
(2) Strong: The video is obviously blurry, details are unclear, the main objects and outlines can be roughly identified, but lack clarity and fineness.
(3) Mild: The blur in the video exists, but it is not too significant and may affect the details.
(4) Slight: The video is basically clear, most objects and details can be clearly identified, with only slight or short-term blur.
(5) Undistorted: The video is very clear, all objects and details can be clearly distinguished, and there is almost no blur phenomenon.

\paragraph{Color dimension:}
(1) Bad: There are obvious defects in color, such as unrealistic colors, hue deviation, extreme exposure or low-light. Or some obvious color errors can be observed, the colors are unnatural, and the viewing experience is seriously affected.
(2) Poor: The color display is inaccurate, but not to an extreme degree. There are some problems with color contrast, saturation, and brightness, but they will not seriously affect the comprehensibility or viewing experience of the content.
(3) Fair: The video color is fair, with slight problems with color contrast, saturation, and brightness.
(4) Good: The color display is basically accurate, and there is no obvious distraction when watching, but the color may not be particularly attractive.
(5) Excellent: The color display is completely accurate, with no visible distortion or offset. The color is attractive.

\paragraph{Temporal dimension:}
(1) Bad: The video frequently has obvious frame rate problems, strobing, jitter or stuttering, resulting in very incoherent content.
(2) Poor: The video has unstable frame rate, occasional frame skipping or slight strobing, jitter or stuttering, which obviously interferes with the coherence of the content and the viewing experience.
(3) Fair: The video occasionally has slight frame rate jumps, strobing, jitter or stuttering, but it does not significantly affect the understanding of the content.
(4) Good: The video maintains a stable frame rate and smooth playback in most cases. In a few cases, there is slight instability, but it does not strongly affect the overall viewing experience.
(5) Excellent: The video has excellent temporal consistency, with almost no frame rate problems, strobing, jitter or stuttering, and the overall viewing experience is smooth and coherent.

\begin{figure*}[t]\centering
\includegraphics[width=1\linewidth]{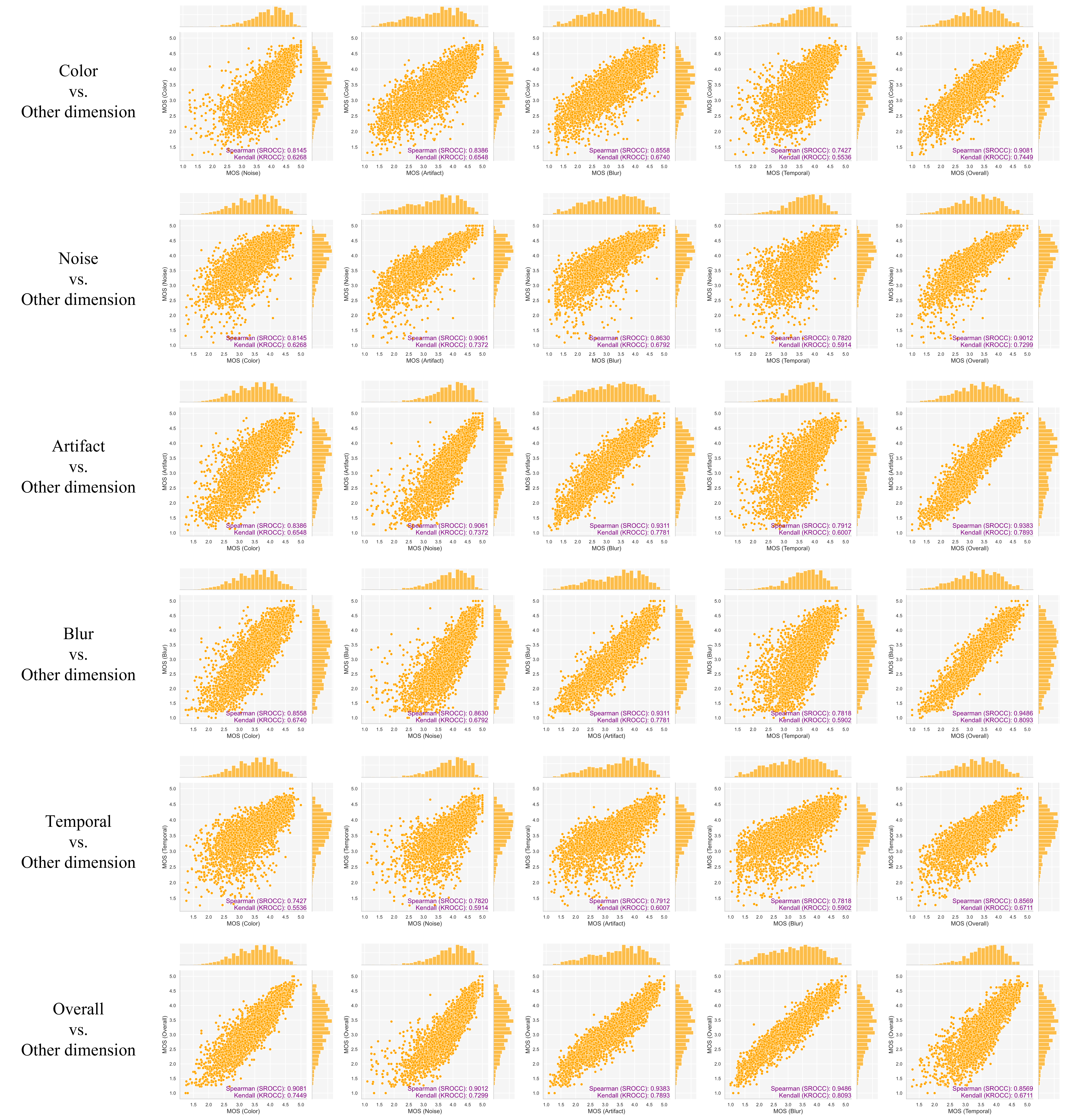}
\vspace{-0.6em}
\caption{Illustration of the MOS correlation between any two dimensions.
}
\label{fig_supp:5}
\end{figure*}

\paragraph{Overall dimension:}
(1) Bad: The video quality is extremely poor, with serious issues in color, noise, artifact, blur, and temporal dimensions, which greatly affect the viewing experience.
(2) Poor: The video quality is poor, with obvious issues in color, noise, artifact, blur, and temporal dimensions, which affect the viewing experience.
(3) Fair: The video quality is fair, with certain problems in color, noise, artifact, blur, and temporal dimensions, which affect the viewing experience to some extent.
(4) Good: The video quality is good, the content is relatively clear, and there is no significant distortion.
(5) Excellent: The video quality is excellent, the content is clear and rich in details, the colors are vivid and realistic, the picture is stable and smooth, and there is no distortion.

For the quality attribute annotation, we ask the subjects to annotate obvious distortions in the video by selecting the options.
If there is no corresponding distortion in the existing options, the subjects can manually enter the distortion.

\subsection{Subjective Data Processing}
For the quality data screening process, we first calculate the kurtosis score of the raw subjective quality ratings for each image to detect it is a Gaussian case or a non-Gaussian case. 
Then, for the Gaussian case, the raw score for an image is considered to be an outlier if it is outside 2 standard deviations (stds) about the mean score of that image; for the non-Gaussian case, it is regarded as an outlier if it is outside $\sqrt{20}$ stds about the mean score of that image. 
A subject is removed if more than 5\% of his/her evaluations are outliers.
For the quality attribute labeling, we choose the options selected by more than half of the subjects as the quality attribute labels.
To generate question-answering (QA) pairs, we query ``yes-or-no'' question for all five dimensions, and additional two ``which exist'' and ``which most affect'' questions over all dimensions.
An overall quality QA pair is also generated.

As a result, we finally obtain numerous fine-grained quality labels for 6104 UGC videos, including 36624 MOSs (6104$\times$6) and 48832 (6104$\times$(5+2+1))) QA pairs.

\subsection{Correlation of Different Dimensions}
To further understand human perceptual quality differences across different dimensions, we also analyze the correlation between any two dimensions in our FineVD database.
As shown in Figure \ref{fig_supp:5}, the correlations are significantly different for different dimension pairs.
First of all, we observe that the color dimension has relatively low correlations with the noise, artifact, blur, and temporal dimensions, but a relatively high correlation with the overall dimension, which manifests that color is a distinct assessment dimension and significantly influence the overall quality.
The noise dimension has a relatively high correlation with the artifact dimension, indicating these two dimensions are related to some extend.
Moreover, the artifact and the blur dimensions exhibit a strong correlation, likely because artifacts often cause blurring.
The temporal dimension has the lowest correlations with all other dimensions, manifesting that the temporal dimension is the most special evaluation dimension.
Finally, the overall dimension has high correlations with almost all dimensions, which further illustrates that the overall quality rating is influenced by all dimensions.

\section{More Details of Our FineVQ Model}

\subsection{Loss Functions}
We use both language loss and L1 loss as the loss functions to optimize the training process.
Specifically, the language loss is used to restrict the FineVQ to produce specific quality attribute, while L1 loss is used to regress the quality scores.
The language loss function can be formulated as:
\begin{equation}
    \mathcal{L}_{\text{language}} = - \frac{1}{N}\sum_{i=1}^{N}\text{log}P(y_\text{label}|y_\text{pred}),
\end{equation}
where $y_\text{pred}$ is the predicted token, $y_\text{label}$ is the ground truth token, $P(y_\text{label}|y_\text{pred})$ indicates the probability, $N$ is the number of tokens.
The L1 loss can be formulated as:
\begin{equation}
    \mathcal{L}_{1} = \frac{1}{N}|q_\text{pred}-q_\text{label}|,
\end{equation}
where $q_\text{pred}$ is the predicted quality score, $q_\text{label}$ is the ground truth quality score, $N$ is the number of videos in a batch.
The overall loss function can be formulated as:
\begin{equation}
    \mathcal{L}\ = \mathcal{L}_{\text{language}} + \mathcal{L}_{1}.
\end{equation}

\subsection{Model and Training Details}
We further describe the dimension of FineVQ in detail. For $E_I$, the output feature dimension is 4096, then two MLPs with the dimension of 4096 is followed to refine the features.
For $E_M$, the output feature dimension is 2304, the two MLPs are followed, which map the feature dimension form 2304 to 4096.
Then, the extracted features are fed into the LLM, whose feature dimension is also 4096.
During training, the image encoder $E_I$, motion encoder $E_M$, text encoder and decoder, and the large language model are frozen, while the projectors and the LoRA weights are trainable.

\section{More Experimental Results}

\subsection{Influence of Extracted Video Frames}

We further conduct an ablation experiment to study the influence of the selected video frame number of image encoder $E_I$ and motion encoder $E_M$, respectively.
As shown in Table \ref{tab:supp_abl}, reducing frame numbers for both image encoder $E_I$ and motion encoder $E_M$ can decrease the performance.
Specifically, comparing the first, second and the last rows in Table \ref{tab:supp_abl}, we can observe that setting the selected frames to 8 for $E_I$ leads to better performance compared to 4 frames and 1 frame.
Moreover, it can be observed that reducing the input frame numbers of $E_M$ also lower the final performance.
Thus, the video frame selection is important in our FineVQ model.

\begin{table}
\renewcommand\arraystretch{1.2}
  \caption{Influence of extracted video frames. $F$ indicates the whole frames.}

  \vspace{-0.5em}
  \resizebox{0.8\linewidth}{!}{
  \begin{tabular}{cc| ccc}
    \toprule
    \multicolumn{2}{c}{Strategy} & \multicolumn{3}{c}{LIVE VQC \cite{sinno2018large}} \\
    \cmidrule(r){1-2} \cmidrule(r){3-5}
    $E_I$ frames  & $E_M$ frames   & SRCC     & PLCC     & KRCC      \\
    \midrule[0.3pt]
    1 & $F$ &  0.8474 & 0.8657 & 0.6743 \\
    4 & $F$ &  0.8609 & 0.8792 & 0.6905 \\
    8 & $F/16$ & 0.8354  & 0.8330 & 0.6444 \\
    8 & $F/4$ &  0.8492 & 0.8525 & 0.6739 \\
    \rowcolor{gray!20} 8 & $F$ &  \bred{0.8951} & \bred{0.8950} & \bred{0.7297} \\
    \bottomrule
  \end{tabular}\label{tab:supp_abl}
  }
  \vspace{-1mm}
  \centering
\end{table}

\subsection{Improvement of Instruction Tuning on the Attribute Prediction Task}
We further compare the performance of FineVQ and our base model, \textit{i.e.,} InternVL2 (\textit{8B}) ~\cite{chen2024internvl} on the \textit{quality attribute prediction} task, and show the results in Table \ref{tab:supp_abl2}.
It can be observed that the established FineVD database and FineVQ model can significantly improve the low-level quality attribute perception ability for the base model, with increasing over 10\% for almost all sub-tasks.

\begin{table}
\setlength{\belowcaptionskip}{-0.01cm}
\centering
\belowrulesep=0pt
\aboverulesep=0pt
\renewcommand\arraystretch{1.2}
\caption{Comparison between FineVQ and the base model InternVL2 (\textit{8B}) ~\cite{chen2024internvl} on our established FineVD database in terms of the \textit{quality attribute prediction task}. The \textit{``yes-or-no''} type represents the judgment on whether the corresponding dimension is degraded. The \textit{``which''} type indicates which distortion exists or has the most impact on the quality of the video.
}
\vspace{-1.5mm}
   \resizebox{\linewidth}{!}{\begin{tabular}{lccccccc}
    \toprule[1pt]
    Question Type&\multicolumn{5}{c}{\textit{Yes-or-no}}&\multicolumn{2}{c}{\textit{Which}}\\
  \cmidrule(lr){2-6} \cmidrule(lr){7-8}
 Model / Attribute&Color&Noise&Artifact&Blur&Temporal&Exist&Most\\
    \midrule
    
    InternVL2 (\textit{8B}) ~\cite{chen2024internvl}&58.46\%&63.58\%&50.69\%&54.33\% &70.28\%&28.25\%&43.21\% \\
    
   \rowcolor{gray!20} \textbf{FineVQ (Ours)} &\bred{73.52\%}&\bred{72.74\%}&\bred{51.87\%}&\bred{64.76\%}&\bred{86.91\%}&\bred{91.93\%}&\bred{65.06\%}\\
   \textit{Improvement} & \blue{15.06\%} & \blue{9.16\%} & \blue{1.18\%} & \blue{10.43\%} & \blue{16.63\%} & \blue{63.68\%} & \blue{21.85\%} \\
    \bottomrule[1pt]
  \end{tabular}}
  \label{tab:supp_abl2}
\end{table}

\end{document}